\documentclass[11pt,draftcls,onecolumn]{IEEEtran}

\usepackage[pdftex]{graphicx}
\graphicspath{{./Figures/}}
\DeclareGraphicsExtensions{.pdf,.jpeg,.png}
\usepackage{amsmath,epsfig}
\usepackage{mathtools}
\usepackage{amssymb}
\usepackage{subfig}
\usepackage{multirow}

\usepackage[utf8]{inputenc} 
\usepackage[T1]{fontenc}    
\usepackage{url}            
\usepackage{booktabs}       
\usepackage{amsfonts}       
\usepackage{nicefrac}       
\usepackage{microtype}      
\usepackage{cite}			

\newcommand{\D}{{\mathcal{D}}}

\usepackage{mathtools}

\usepackage[usenames,dvipsnames]{color}

\newcommand{\errata}[1]{\textcolor{black}{#1}}

\begin{document}

\title{Learning graphs from data: \\A signal representation perspective}
\author{Xiaowen Dong*, Dorina Thanou*, Michael Rabbat, and Pascal Frossard
\thanks{*Authors contributed equally.}}
\maketitle

The construction of a meaningful graph topology plays a crucial role in the effective representation, processing, analysis and visualization of structured data. When a natural choice of the graph is not readily available from the data sets, it is thus desirable to infer or learn a graph topology from the data. In this tutorial overview, we survey solutions to the problem of graph learning, including classical viewpoints from statistics and physics, and more recent approaches that adopt a graph signal processing (GSP) perspective. We further emphasize the conceptual similarities and differences between classical and GSP-based graph inference methods, and highlight the potential advantage of the latter in a number of theoretical and practical scenarios. We conclude with several open issues and challenges that are keys to the design of future signal processing and machine learning algorithms for learning graphs from data.

\section{Introduction}
\label{sec:intro}

Modern data analysis and processing tasks typically involve large sets of structured data, where the structure carries critical information about the nature of the data. One can find numerous examples of such data sets in a wide diversity of application domains, including transportation networks, social networks, computer networks, and brain networks.
Typically, graphs are used as mathematical tools to describe the structure of such data. They provide a flexible way of representing relationship between data entities. Numerous signal processing and machine learning algorithms have been introduced in the past decade for analyzing structured data on \emph{a priori} known graphs \cite{Zhu05,Fortunato10,Shuman13}. However, there are often settings where the graph is not readily available, and the structure of the data has to be estimated in order to permit effective {representation, processing, analysis or visualization of graph data.}  In this case, a crucial task is to infer a graph topology that describes the characteristics of the {data observations}, hence capturing the underlying relationship between these entities.

\begin{figure}[t]
      \centering
        \includegraphics[width=16cm]{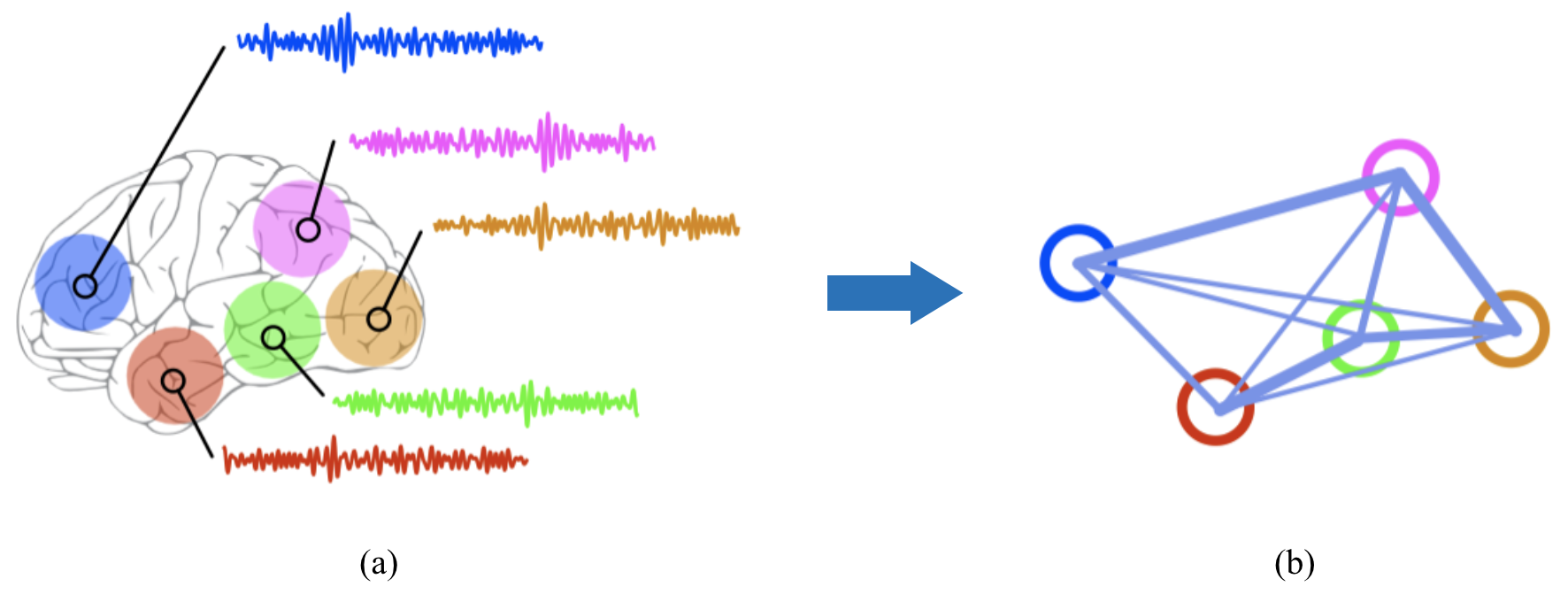}		\label{brain-example}
        \caption{{Inferring functional connectivity between different regions of the brain. (a) BOLD time series recorded in different regions of the brain. (b) A functional connectivity graph where the vertices represent the brain regions and the edges (with thicker bars indicating heavier weights) represent the strength of functional connections between these regions. Figure adapted from \cite{Richiardi13} with permission.}}
        \label{fig:brain-example}
\end{figure}

{Consider an example in brain signal analysis. Suppose we are given blood-oxygen-level-dependent (BOLD) signals, which are time series extracted from functional magnetic resonance imaging (fMRI) data that reflect the activities of different regions of the brain. An area of significant interest in neuroscience is to infer functional connectivity, i.e., capture relationship between brain regions which correlate or synchronize given a certain condition of a patient, which may help reveal underpinnings of some neurodegenerative diseases (see Fig.~\ref{fig:brain-example} for an illustration). This leads to the problem of inferring a graph structure given the multivariate BOLD time series data.}

{Formally, the problem of graph learning is the following: given $M$ observations on $N$ variables or data entities, represented in a data matrix $\mathbf{X} \in \mathbb{R}^{N \times M}$, and given some prior knowledge (e.g., distribution, data model, etc) about the data, we would like to build or infer relationship between these variables that take the form of a graph $\mathcal{G}$. As a result, each column of the data matrix $\mathbf{X}$ becomes a graph signal defined on the node set of the estimated graph, and the observations can be represented as $\mathbf{X}=\mathcal{F}(\mathcal{G})$, where $\mathcal{F}$ represents a certain generative process or function on the graph.}

{The graph learning problem is an important one because: 1) a graph may capture the actual geometry of structured data, which is essential to efficient processing, analysis and visualization; 2) learning relationship between data entities benefits numerous application domains, such as understanding functional connectivity between brain regions or behavioral influence between a group of people; 3) the inferred graph can help in predicting data evolution in the future.}

Generally speaking, inferring graph topologies from observations is an ill-posed problem, and there are many ways of associating a topology with the observed data samples. {Some of the most straightforward methods include computing sample correlation, or using a similarity function, e.g., a Gaussian RBF kernel function, to quantify the similarity between data samples. These methods are based purely on observations without any explicit prior or model of the data, hence they may be sensitive to noise and have difficulty in tuning the hyper-parameters.}
A meaningful data model or accurate prior may, however, guide the graph inference process and lead to a graph topology that better reveals the intrinsic relationship among the data entities.
Therefore, a main challenge in this problem is to define such a model for the generative process or function $\mathcal{F}$,
such that it captures the relationship between the observed data $\mathbf{X}$ and the learned graph topology $\mathcal{G}$. 
Naturally, such models often correspond to specific criteria for describing or estimating structures between the data samples, e.g., models that put a smoothness assumption on the data, or that represent an information diffusion process on the graph.

Historically, there have been two general approaches to learning graphs from data, one based on statistical models and one based on physically-motivated models.
From the statistical perspective, 
$\mathcal{F}(\mathcal{G})$ is modeled as {a function that draws a realization from} a probability distribution over {the variables} that is determined by the structure of $\mathcal{G}$. One prominent example is found in probabilistic graphical models \cite{Koller09},
where the graph structure encodes conditional independence relationship among random variables that are represented by the vertices. Therefore, learning the graph structure is equivalent to learning a factorization of a \emph{joint probability distribution} of these random variables. 
{Typical application domains include inferring interactions between genes using gene expression profiles, and relationship between politicians given their voting behavior \cite{Banerjee08}.}

\begin{figure}[t]
      \centering
        \includegraphics[width=12cm]{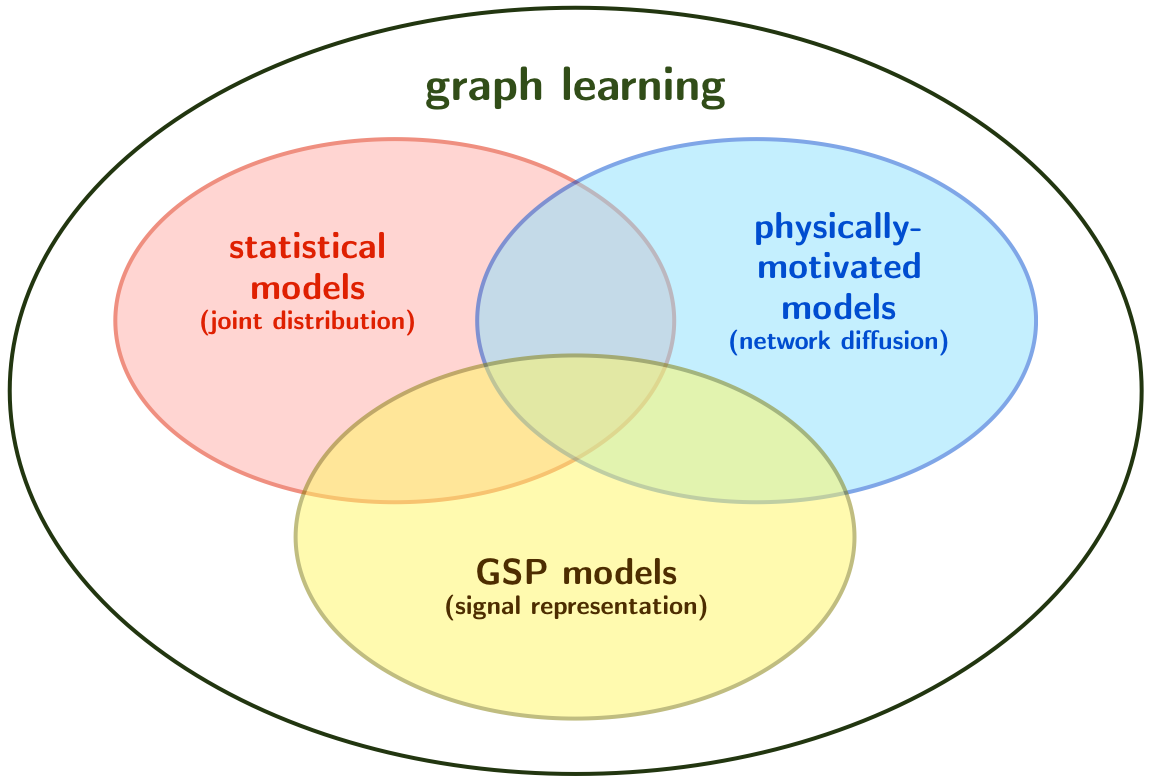}		\label{approaches}
        \caption{A broad categorization of different approaches to the problem of graph learning.}
        \label{fig:approaches}
\end{figure}

For physically-motivated models, $\mathcal{F}(\mathcal{G})$ is defined based on the assumption of an underlying physical phenomenon or process on the graph. One popular process is \emph{network diffusion or cascades} \cite{GomezRodriguez_2010,Myers2010,GomezRodriguez2014,Nan2012}, where $\mathcal{F}(\mathcal{G})$ dictates the diffusion behavior on $\mathcal{G}$ that leads to the observation of $\mathbf{X}$, possibly at different time steps.
In this case, the problem is equivalent to learning a graph structure on which the generative process of the observed signals may be {explained}.
Practical applications include understanding information flowing over a network of online media sources \cite{GomezRodriguez_2010} or observing epidemics spreading over a network of human interactions \cite{Groendyke11}, {given the state of exposure or infection at certain time steps.}

The fast growing field of graph signal processing \cite{Shuman13,Sandryhaila13} offers a new perspective to the problem of graph learning. In this setting, the columns of the observation matrix $\mathbf{X}$ are {explicitly} considered as signals that are defined on the vertex set of a weighted graph $\mathcal{G}$. The learning {problem} can then be cast as one of learning a graph $\mathcal{G}$ such that {$\mathcal{F}(\mathcal{G})$ permits to make certain properties or characteristics of the observations $\mathbf{X}$ explicit, e.g., smoothness with respect to $\mathcal{G}$ or sparsity in a basis related to $\mathcal{G}$.} 
{This \emph{signal representation} perspective is particularly interesting as it puts a strong and {explicit} emphasis on the relationship between the signal representation and the graph topology, where $\mathcal{F}(\mathcal{G})$ often comes with an interpretation of frequency-domain analysis or filtering operation of signals on the graph. For example, it is typical to adopt the eigenvectors of the graph Laplacian matrix associated with $\mathcal{G}$ as a surrogate for the Fourier basis for signals supported on $\mathcal{G}$ \cite{Shuman13,Ortega18}; we go deeper into the details of this view in Sec.~\ref{sec:gsp}.}

One common representation of interest is a smooth representation in which $\mathbf{X}$ has a slow variation on $\mathcal{G}$, which can be interpreted as $\mathbf{X}$ mainly consisting of low frequency components in the graph spectral domain. 
{Such Fourier-like analysis on the graph leads to {novel graph inference methods compared to} approaches rooted in statistics or physics; more importantly,}
it offers the {opportunity} to represent $\mathbf{X}$ in terms of its behavior in the graph spectral domain,
which makes it possible to capture complex and non-typical behavior of graph signals that cannot be explicitly handled by classical tools,
{for example bandlimited signals on graphs.}
Therefore, 
{given potentially more accurate assumptions underlying the GSP models,}
the inference of $\mathcal{G}$ given a specifically designed $\mathcal{F}$ may better reveal the intrinsic relationship between the data entities and benefit subsequent data processing applications.
{Conceptually, as illustrated in Fig.~\ref{fig:approaches}, GSP-based graph learning approaches can thus be considered as a new family of methods that {have close connections} with classical methods while also offering certain unique advantages {in graph inference.}}

In this tutorial overview, we first review well-established solutions to the problem of graph learning {that adopt} a statistics or a physics perspective.
Next, we survey a series of recent GSP-based approaches and show how signal processing tools and concepts can be utilized to provide novel solutions to the {graph learning} problem.
Finally, we showcase applications of GSP-based methods in a number of domains and conclude with open questions and challenges that are central to the design of future signal processing and machine learning algorithms for learning graphs from data.

\section{Literature review}
\label{sec:literature}
The recent availability of a large amount of data collected in a variety of application domains leads to an increasing interest in estimating the structure, often encoded in the form of a network or a graph, that underlies the data.
Two general approaches have been proposed in the literature, one based on statistical models and the other based on physically-motivated models. We provide a detailed review of these two approaches next.

\subsection{Statistical models}
\label{sec:statistical}
The general philosophy behind the statistical view is that there exists a graph $\mathcal{G}$ whose structure determines the joint probability distribution of the observations on the data entities, i.e., {columns of the data matrix $\mathbf{X}$}.
In this case, the function $\mathcal{F}(\mathcal{G})$ in our problem formulation is one that draws a collection of realizations, i.e., the columns of $\mathbf{X}$, from the distribution governed by $\mathcal{G}$.
Such models are known as probabilistic graphical models \cite{Koller09,Meinshausen06,Banerjee08,Friedman08,Hsieh11},
{where the edges (or lack thereof) in the graph encode conditional independence relationship among the random variables represented by the vertices.} 

There are two main types of graphical models: 1) undirected graphical models, also known as Markov random fields (MRFs), in which local neighborhoods of the graph capture the independence structure of the variables; and 2) directed graphical models, also known as Bayesian networks or belief networks (BNs), which have a more complicated notion of independence by taking into account the direction of edges.
Both MRFs and BNs have their respective advantages and disadvantages. In this {section}, we focus primarily on the approaches for learning MRFs, {which admit a simpler representation of conditional independence and also have connections to GSP-based methods, as we will see later.}
Readers who are interested in the comparison between MRFs and BNs as well as approaches for learning BNs are referred to \cite{Koller09,Heckerman95}.

An MRF with respect to a graph $\mathcal{G}=\{\mathcal{V},\mathcal{E}\}$, where $\mathcal{V}$ and $\mathcal{E}$ denote the vertex and edge set, respectively, is a set of random variables $\mathbf{x} = \{x_i : v_i \in \mathcal{V}\}$ that satisfy a Markov property. We are particularly interested in the pairwise Markov property:
\begin{equation}
(v_i,v_j) \notin \mathcal{E} \Leftrightarrow p(x_i | x_j, \mathbf{x} \setminus \{x_i, x_j\}) = p(x_i | \mathbf{x} \setminus \{x_i, x_j\}).
\label{eq:markov}
\end{equation}
Eq.~(\ref{eq:markov}) states that two variables $x_i$ and $x_j$ are conditionally independent given the rest if there is no edge between the corresponding vertices $v_i$ and $v_j$ in the graph.
Suppose we have $N$ random variables, then this condition holds for the {exponential family of distributions} with a parameter matrix $\mathbf{\Theta} \in \mathbb{R}^{N \times N}$:
\begin{equation}
p(\mathbf{x}|\mathbf{\Theta}) = \frac{1}{Z(\mathbf{\Theta})} \text{exp} \left( \sum_{v_i \in \mathcal{V}} \theta_{ii}x_i^2 + \sum_{(v_i,v_j) \in \mathcal{E}} \theta_{ij}x_i x_j \right),
\end{equation}
where $\theta_{ij}$ represents the $ij$-th entry of $\mathbf{\Theta}$, and $Z(\mathbf{\Theta})$ is a normalization constant.

Pairwise MRFs consist of two main classes: 1) Gaussian graphical models or Gaussian MRFs (GMRFs), in which the variables are continuous; 2) discrete MRFs, in which the variables are discrete. In the case of a (zero-mean) GMRF, the joint probability can be written as follows:
\begin{equation}
p(\mathbf{x}|\mathbf{\Theta}) = \frac{|\mathbf{\Theta}|^{1/2}}{(2 \pi)^{N/2}} \text{exp} \big( -\frac{1}{2} \mathbf{x}^T \mathbf{\Theta} \mathbf{x} \big),
\end{equation}
where $\mathbf{\Theta}$ is the inverse covariance or \emph{precision} matrix.
In this context, learning the graph structure boils down to learning the matrix $\mathbf{\Theta}$ that encodes pairwise conditional independence between the variables. It is common to assume, or take as a prior, that $\mathbf{\Theta}$ is sparse because: 1) real world interactions are typically local; 2) the sparsity assumption makes learning computationally more tractable. In what follows, we review some key developments in learning Gaussian and discrete MRFs.

\begin{figure}[t]
      \centering
        \subfloat[] {\includegraphics[width=3cm]{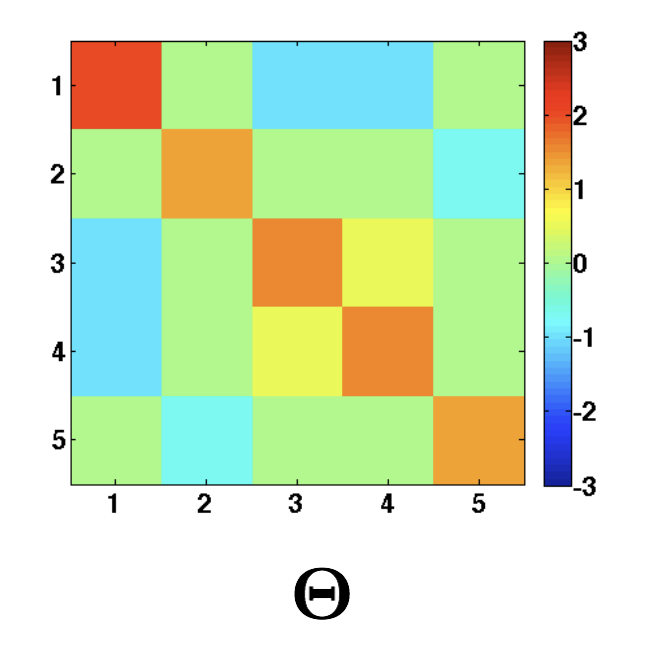}}~
        \subfloat[] {\includegraphics[width=4.86cm]{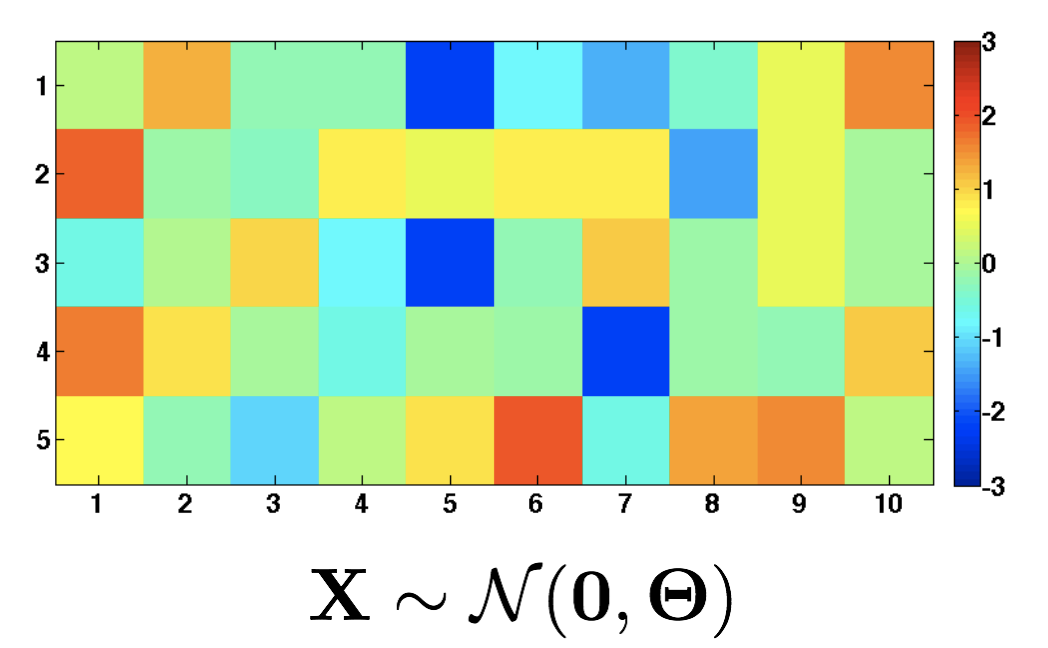}}~
        \subfloat[] {\includegraphics[width=3cm]{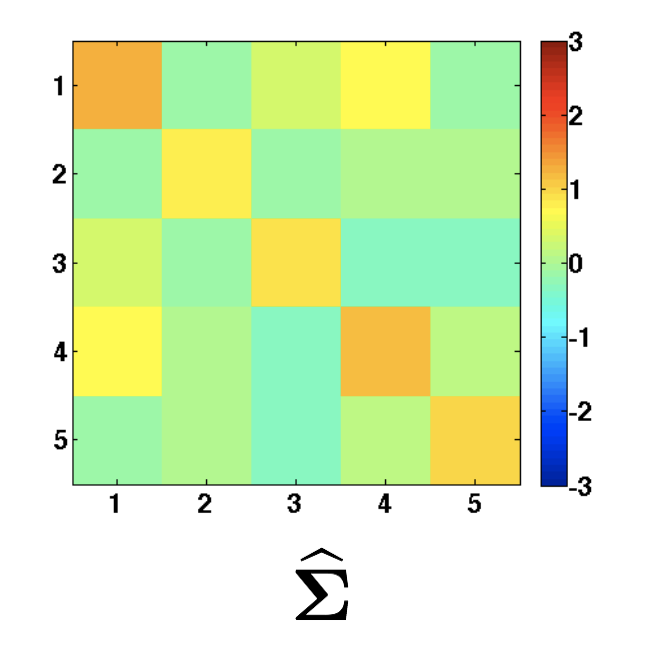}}~
        \subfloat[] {\includegraphics[width=3cm]{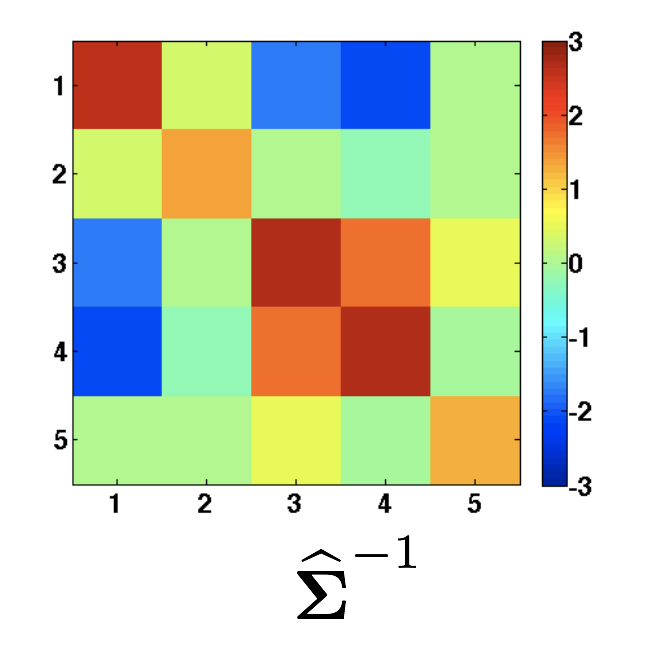}}
        \caption{(a) A groundtruth precision $\mathbf{\Theta}$. (b) An observation matrix $\mathbf{X}$ drawn from a multivariate Gaussian distribution with $\mathbf{\Theta}$. (c) The sample covariance $\widehat{\boldsymbol{\Sigma}}$. (d) The inverse of the sample covariance $\widehat{\boldsymbol{\Sigma}}$.}
        \label{fig:invcov}
\end{figure}

For learning GMRFs, one of the first approaches is suggested in \cite{Dempster72}, where the author has proposed to learn $\mathbf{\Theta}$ by sequentially pruning the smallest elements in the inverse of the sample covariance matrix $\widehat{\boldsymbol{\Sigma}} = \frac{1}{M-1}\mathbf{X} \mathbf{X}^T$ (see Fig.~\ref{fig:invcov}). 
Although it is based on a simple and effective rule, {this method does not perform well when the sample covariance is not a good approximation of the ``true'' covariance, often due to a small number of samples.}
{In fact, the method cannot even be applied} when the sample size is smaller than the number of variables, in which case the sample covariance matrix is not invertible.

\begin{figure}[t]
      \centering
        \subfloat[] {\includegraphics[width=4cm]{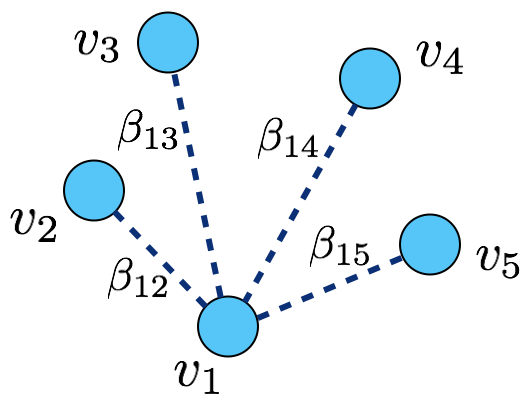}}~
        \subfloat[] {\includegraphics[width=6cm]{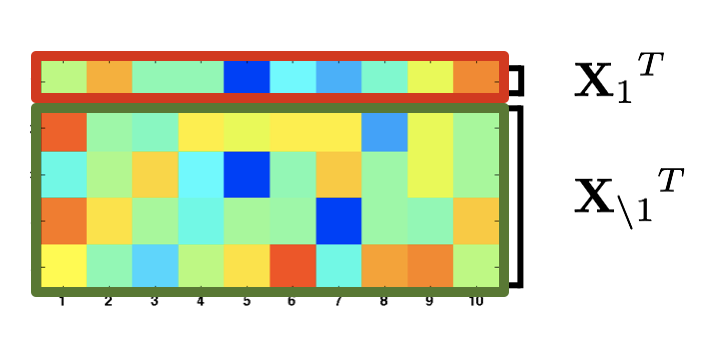}}~
        \subfloat[] {\includegraphics[width=6cm]{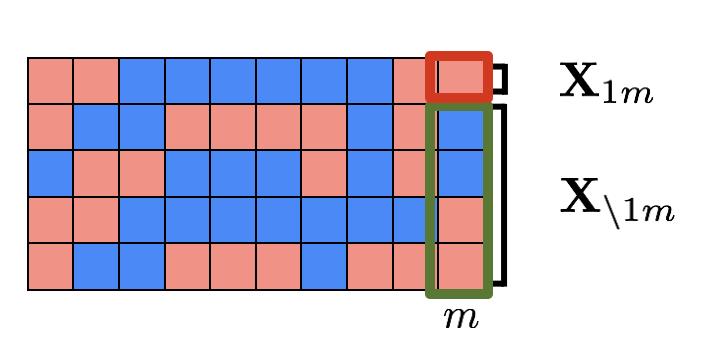}}
        \caption{(a) Learning graphical models by neighborhood selection. (b) Neighborhood selection via the Lasso regression for Gaussian MRFs. (c) Neighborhood selection via logistic regression for discrete MRFs.}
        \label{fig:ns}
\end{figure}

{Since a graph is a representation of pairwise relationship, it is clear that learning a graph is equivalent to learning a neighborhood for each vertex, i.e., the other vertices to which it is connected. In this case, it is natural to assume that the observation at a particular vertex may be represented by observations at the neighboring vertices.
Based on this assumption,} the authors in \cite{Meinshausen06} have proposed to approximate the observation at each variable as a sparse linear combination of the observations at other variables. For a variable $x_1$, for instance, this approximation leads to a Lasso regression problem \cite{Tibshirani96} of the form:
\begin{equation}
\underset{\boldsymbol{\beta}_1}{\text{min}}~ || \mathbf{X}_1 - \mathbf{X}_{\backslash 1}\boldsymbol{\beta}_1||_2^2 + \lambda || \boldsymbol{\beta}_1||_1,
\label{eq:ns}
\end{equation}
where $\mathbf{X}_1$ and $\mathbf{X}_{\backslash 1}$ represent the observations on the variable $x_1$ (i.e., transpose of the first row of $\mathbf{X}$) and the rest of the variables, respectively, and $\boldsymbol{\beta}_1 \in \mathbb{R}^{N-1}$ is a vector of coefficients for $x_1$ (see Fig.~\ref{fig:ns}(a)-(b)). 
In Eq.~(\ref{eq:ns}), the first term can be interpreted as 
\errata{the negative} local log-likelihood of $\boldsymbol{\beta}_1$ and the $L^1$ penalty is added to enforce its sparsity, with a regularization parameter $\lambda$ balancing the two terms. The same procedure is then repeated for all the variables (or vertices). 
\errata{Finally, a connection between a pair of vertices $v_i$ and $v_j$ is established if either of $\beta_{ij}$ and $\beta_{ji}$ is nonzero, or both (notice that it should not be interpreted that $\beta_{ij}$ and $\beta_{ji}$ are directly related to the corresponding entries in the precision matrix $\boldsymbol{\Theta}$). This \emph{neighborhood selection} approach using the Lasso is intuitive with certain theoretical guarantees \cite{Meinshausen06}; however, it does not involve solving an optimization problem whose objective is an explicit function of $\boldsymbol{\Theta}$.}

{Instead of per-node neighborhood selection, the works in \cite{Yuan06,Banerjee08,Friedman08} have proposed a popular method for estimating an inverse covariance or precision matrix at once, which is based on maximum likelihood estimation.} Specifically, the so-called \emph{graphical Lasso} method aims to solve the following problem:
\begin{equation}
\underset{\boldsymbol{\Theta}}{\text{max}}~\text{log}~\text{det} \boldsymbol{\Theta} - \mathrm{tr}(\widehat{\boldsymbol{\Sigma}}\boldsymbol{\Theta}) - \rho ||\boldsymbol{\Theta}||_1,
\label{eq:gLasso}
\end{equation}
where $\widehat{\boldsymbol{\Sigma}}$ is the sample covariance matrix\footnote{\errata{In the graphical Lasso formulation, the sample covariance is computed as $\widehat{\boldsymbol{\Sigma}} = \frac{1}{M}\mathbf{X} \mathbf{X}^T$.}}, 
and $\text{det}(\cdot)$ and $\mathrm{tr}(\cdot)$ represent the determinant and trace operators, respectively. The first two terms together can be interpreted as the log-likelihood under a GMRF and {the entry-wise $L^1$ norm of $\boldsymbol{\Theta}$} is added to enforce sparsity of the connections with a regularization parameter $\rho$. The main difference between this approach and the neighborhood selection method of \cite{Meinshausen06} is that the optimization in the latter is decoupled for each vertex, while the one in graphical Lasso is coupled, which can be essential for stability under noise.
Although the problem of Eq.~(\ref{eq:gLasso}) is convex, log-determinant programs are in general computationally demanding. Nevertheless, a number of efficient approaches have been proposed specifically for the graphical Lasso. For example, the work in \cite{Hsieh11} proposes a quadratic approximation of the Gaussian negative log-likelihood that can significantly speed up optimization.

Unlike the GMRFs, variables in discrete MRFs take discrete values. One popular example is the binary Ising model \cite{Cipra87}. Various learning methods may be applied in such cases, and one notable example is the approach proposed in \cite{Ravikumar10}, based on the idea of neighborhood selection similar to that in \cite{Meinshausen06}. Specifically, given the exponential family distribution introduced before, it is easy to verify that the conditional probability of one variable given the rest, e.g., $p(\mathbf{X}_{1m}|\mathbf{X}_{{\backslash 1}m})$ for variable $x_1$ where $\mathbf{X}_{1m}$ and $\mathbf{X}_{{\backslash 1}m}$ respectively represent the first entry and the rest of the $m$-th column of $\mathbf{X}$ (see Fig.~\ref{fig:ns}(c)), follows the form of a logistic function. Therefore, $x_1$ can be considered as the dependent variable in a logistic regression where all the other variables serve as independent variables. To learn sparse connections within the neighborhood of this vertex, the authors of \cite{Ravikumar10} have proposed to solve an $L^1$-regularized logistic regression:
\begin{equation}
\errata{\underset{\boldsymbol{\beta}_1}{\text{max}}~ \sum_{m=1}^M \text{log}~ p_{\boldsymbol{\beta}_1} (\mathbf{X}_{1m}|\mathbf{X}_{{\backslash 1}m}) - \lambda || \boldsymbol{\beta}_1||_1.}
\end{equation}
The same procedure is then repeated for the rest of the vertices to compute the final connection matrix, similar to that in \cite{Meinshausen06}.

Most previous approaches for learning GMRFs recover a precision matrix with both positive and negative entries. A positive off-diagonal entry in the precision matrix implies a negative partial correlation between the two random variables, which is difficult to interpret in some contexts, such as road traffic networks. For such application settings, it is therefore desirable to learn a graph topology with non-negative weights. To this end, the authors in \cite{Slawski15} have proposed to select the precision matrix from the family of the so-called M-matrices \cite{Poole74}, which are symmetric and positive definite matrices with non-positive off-diagonal entries, leading to the \emph{attractive} GMRFs.
Since the graph Laplacian matrix $\mathbf{L}$ is a (singular) M-matrix that {uniquely determines the adjacency matrix $\mathbf{W}$}, {it is a popular modeling choice and numerous papers have focused on learning $\mathbf{L}$ as a specific instance of the precision matrices.}

One notable example is the work in \cite{Lake10}, which adapts the graphical Lasso formulation of Eq.~(\ref{eq:gLasso}) and proposes to solve the following problem{\footnote{{The exact formulation of the optimization problem in \cite{Lake10} is in a slightly different but equivalent form, due to the following relationship:
$||\boldsymbol{\Theta}||_1 = ||\mathbf{L}||_1 + \frac{1}{\sigma^2}N = 2||\mathbf{W}||_1 + \frac{1}{\sigma^2}N.$
We therefore choose the formulation in Eq.~(\ref{eq:lake}) as it illustrates the connection with the graphical Lasso formulation in a straightforward way.}}}:
\begin{equation}
\begin{split}
\underset{\boldsymbol{\Theta},~\sigma^2}{\mbox{maximize}} ~~~ & \text{log}~\text{det} \boldsymbol{\Theta} - \mathrm{tr}(\frac{1}{M}\mathbf{X} \mathbf{X}^T \boldsymbol{\Theta}) - \rho ||\boldsymbol{\Theta}||_1, \\
\mbox{subject to} ~~~ & \mathbf{\Theta} = \mathbf{L} + \frac{1}{\sigma^2} \mathbf{I},~\mathbf{L} \in \mathcal{L},
\end{split}
\label{eq:lake}
\end{equation}
\noindent where $\mathbf{I}$ is the identity matrix, ${\sigma^2>0}$ is the a priori feature variance, {$\mathcal{L}$ is the set of valid graph Laplacian matrices, and $||\cdot||_1$ represents the entry-wise $L^1$ norm.} In Eq.~(\ref{eq:lake}), the precision matrix $\boldsymbol{\Theta}$ is modeled as a regularized graph Laplacian matrix (hence full-rank). By solving for it, the authors obtain the graph Laplacian matrix, or in other words, an adjacency matrix with non-negative weights.

Notice that the trace term in Eq.~(\ref{eq:lake}) includes the so-called Laplacian quadratic form $\mathbf{X}^T \mathbf{L} \mathbf{X}$, which measures the smoothness of the data on the graph and has also been used in other approaches that are not necessarily developed from the viewpoint of inverse covariance estimation. For instance, the works in \cite{Daitch09} and \cite{Hu15} have proposed to learn the graph by minimizing quadratic forms that involve powers of the graph Laplacian matrix $\mathbf{L}$. When the power of the Laplacian is set to two, this is equivalent to the locally linear embedding criterion proposed in \cite{Roweis00} for nonlinear dimensionality reduction. 
{As we shall see in the following section, the criterion of signal smoothness has also been adopted in one of the GSP models for graph inference.}

\subsection{Physically-motivated models}
\label{sec:physics}
{While the above methods mostly exploit statistical properties for graph inference, in particular the conditional independence structure between random variables,
another family of approaches tackles the problem by taking a physically-motivated perspective.}
{In this case, the observations $\mathbf{X}$ are considered as outcomes of some physical phenomena on the graph, {specified by the function $\mathcal{F}(\mathcal{G})$,} and the inference problem consists in capturing the structure inherent to the physics of the observed data.} Two examples of such methods are 1) network tomography, where the physical process models data actually transmitted in a communication network, and 2) epidemic or information propagation models, where the physical process represents a disease spreading over a contact network or a meme spreading over social media.

The field of \emph{network tomography} broadly concerns methods for inferring properties of networks from indirect observations \cite{Castro2004network}. It is most commonly used in the context of telecommunication networks, where the information to be inferred may include {the network routes}, or the properties such as available bandwidth or reliability of each link in the network.
{For example, end-to-end measurements are acquired by sending a sequence of packets from one source to many destinations, and sequences of received packets are used to infer the internal network topology.}
The seminal work on this problem aimed to infer the routing tree from one source to multiple destinations \cite{Ratnasamy1999inference}. Subsequent work considered interleaving measurements from multiple sources to the same destinations simultaneously to infer general topologies \cite{Rabbat2006multiple}. {These methods can be interpreted as choosing the function $\mathcal{F}(\mathcal{G})$ in our formulation as one that measures network responses by exhaustively
sending probes between all possible pairs of end-hosts.} 
Consequently, this may impose a significant amount of measurement traffic on the network. In order to reduce this traffic, approaches based on active sampling have also been proposed \cite{Sattari2014active}.

{\emph{Information propagation} models have been applied to infer latent biological, social and financial networks based on observations of epidemics, memes, or other signals diffusing over them (e.g., \cite{GomezRodriguez_2010,Myers2010,GomezRodriguez2014,Nan2012}).
For simplicity and consistency, in our discussion, we adopt the terminology of epidemiology. This type of models is characterized by three main components: (a) the \emph{nodes}, (b) an \emph{infection process} (i.e., the change in the state of the node that is transferred by neighboring nodes in the network), and (c) the \emph{causality} (i.e., the underlying graph structure based on which the infection is propagated). 
Given a known graph structure, epidemic processes over graphs have been well-studied through popular models in which nodes may be susceptible, infected, and possibly recovered \cite{PastorSatorras15}.
On the other hand, when the structure is not known beforehand, it may be inferred by considering the propagation of contagions over the edges of an unknown network, 
given usually only the time steps when nodes became infected.}

\begin{figure}[t]
      \centering
        \subfloat[] {\includegraphics[width=8cm]{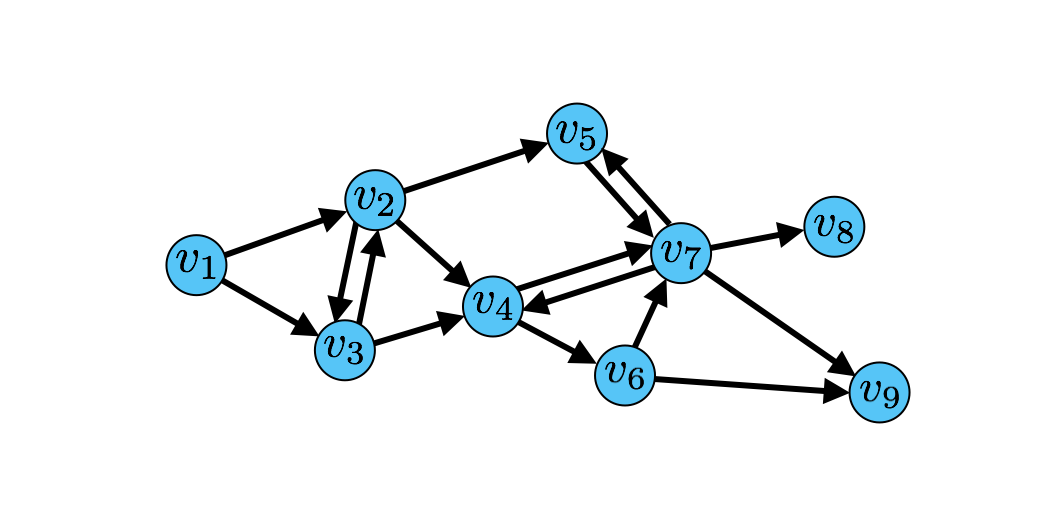}}~
        \subfloat[] {\includegraphics[width=8cm]{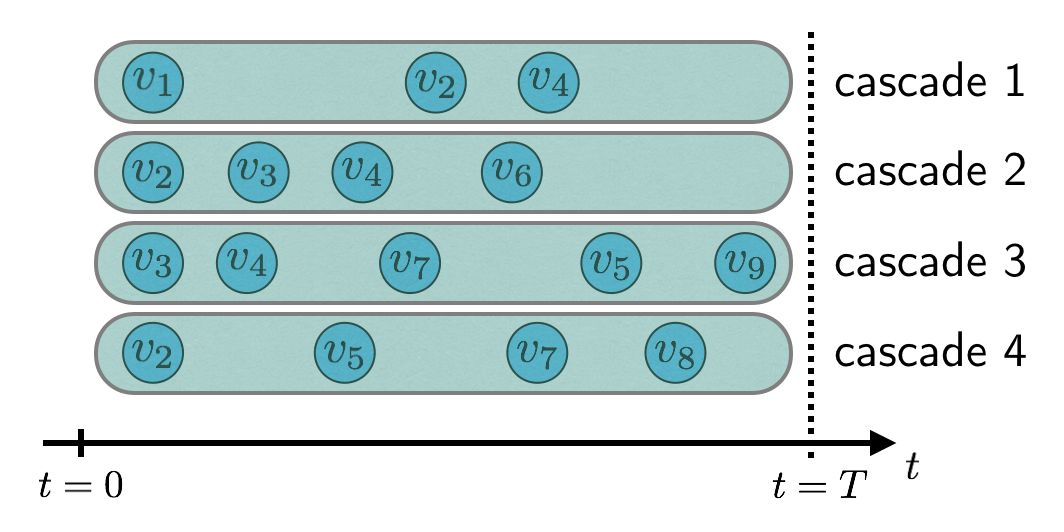}}
        \caption{(a) A graph with directed edges indicating possible directions of spreading. (b) Observations of cascades spreading over the graph. We observe the times when nodes became infected (i.e., the cascade reached a node) but do not observe from which neighbor it was infected. Figure inspired by the one in \cite{GomezRodriguez2014}.}
        \label{fig:cascade}
\end{figure}

{A (fully-observed) cascade may be represented by the sequence of triples $\{(v'_{p}, v_{p}, t_p)\}_{p = 0}^P$, where $P \le N$, representing that node $v'_{p}$ infected its neighbor $v_{p}$ at time $t_p$. In many applications, one may observe when a node becomes infected, but not which neighbor infected it (see Fig.~\ref{fig:cascade} for an illustration). Then, the task is to recover a graph $\mathcal{G}$ given the (partial) observations $\{(v_{p}, t_p)\}_{p =0}^P$, usually for a number of such cascades.
In this case, the set of nodes is given and the goal is to recover the edge structure. The common convention is to shift the infection times so that the initial infection in each cascade always occurs at time $t_0 = 0$. Equivalently, let $\mathbf{x}$ denote a length-$N$ vector where $x_i$ is the time when $v_i$ is infected, using the convention that $x_i = \infty$ if $v_i$ is not infected in this cascade. The observations from $M$ cascades can then be represented in a $N$-by-$M$ matrix $\mathbf{X} = \mathcal{F}(\mathcal{G})$.}

{Methods for inferring networks from information cascades can be generally divided into two main categories depending on whether they are based on homogeneous or heterogeneous models. Methods based on \emph{homogeneous} models assume that cascades propagate in a statistically identical manner across all edges. For example, one model treats entries $w_{ij}$ of the (unknown) adjacency matrix as representing the conditional probability that $v_i$ infects $v_j$ given $v_i$ is infected~\cite{Myers2010}. In addition, a transmission time model $h(t)$ is assumed known such that the likelihood that $v_i$ infects $v_j$ at time $x_j$ given that $v_i$ was infected at time $x_i < x_j$ is:
\begin{equation}
p(x_j | x_i, w_{ij}) = h(x_j - x_i) w_{ij}.
\end{equation}
Here, $h(t)$ is taken to be zero for $t < 0$, and typically $h(t)$ also decays to zero as $t \rightarrow \infty$.}

{Assuming that the function $h(t)$ is given, the inference problem reduces to finding the conditional probabilities $w_{ij}$. 
Given the set of nodes infected as well as the time of infection in each observed cascade, and assuming that cascades are independent and identically distributed, the likelihood of a graph with adjacency matrix $\mathbf{W}$ (with $w_{ij}$ being the $ij$-th entry) is derived explicitly in \cite{Myers2010}, and it is further shown that maximizing this likelihood can be recast as an equivalent geometric program, so that convex optimization techniques can be applied to the problem of inferring $\mathbf{W}$.} 

{A similar model is considered in~\cite{GomezRodriguez_2010}, in which the conditional transmission probabilities are taken to be the same on all edges, i.e., 
$w_{ij} = \beta \cdot \mathbf{1}\{(v_i,v_j) \in \mathcal{E}\}$ where $\mathbf{1}\{\cdot \}$ is an indicator function, for a given constant $\beta \in (0,1)$.
The task therefore reduces to determining where there are edges, which is a discrete optimization problem. The maximum likelihood objective is shown to be submodular in~\cite{GomezRodriguez_2010}, and an edge selection scheme based on greedy optimization obtains the optimal likelihood up to a constant factor. Clearly, the main drawbacks of homogeneous methods are the strong underlying assumption that cascades propagate in an identical manner across all edges in the network.}

\newcommand{\sur}{\mathop{\operatorname{Sur}}}
\newcommand{\haz}{\mathop{\operatorname{Haz}}}
{Methods based on \emph{heterogeneous} models relax these requirements and allow for cascades to propagate at different rates across different edges. The \textsc{NetRate} algorithm~\cite{GomezRodriguez2014} is a prototypical example of this category, in which one assumes a parametric form for the edge conditional likelihood $p(x_j | x_i, w_{ij})$. For example, in an exponential model, $p(x_j | x_i, w_{ij}) = w_{ij} e^{-w_{ij} (x_j - x_i)} \cdot \mathbf{1}\{x_j > x_i\}$. If we write $P(x_j | x_i, w_{ij}) = \int_{x_i}^{x_j} p(t | x_i, w_{ij}) \;dt$ for the cumulative density function, then the \emph{survival function}
\begin{equation}
\sur(x_j | x_i, w_{ij}) := 1 - P(x_j | x_i, w_{ij})
\end{equation}
is the probability that $v_j$ is not infected by $v_i$ by time $x_j$ given that $v_i$ was infected at time $x_i$. Furthermore, the \emph{hazard function}
\begin{equation}
\haz(x_j | x_i, w_{ij}) := \frac{p(x_j | x_i, w_{ij})}{\sur(x_j | x_i, w_{ij})}
\end{equation}
is the instantaneous probability, at time $x_j$, that $v_j$ is infected by $v_i$ given that $v_i$ was infected at time $x_i$.}

{
With this notation, the likelihood of a given cascade observation $\mathbf{x}$ that is observed up to time $T = \max\{x_v < \infty \colon v \in \mathcal{V}\}$ is~\cite{GomezRodriguez2014}:
\begin{equation}
\begin{split}
p(\mathbf{x} | \mathbf{W}) &= \prod_{i : x_i \le T} \prod_{j : x_j > T} \sur(T | x_i, w_{ij}) \\
&\quad \times \prod_{k: x_k < x_i} \sur(x_i | x_k, w_{ki}) \sum_{l : x_l < x_i} \haz(x_i | x_l, w_{li}).
\end{split}
\end{equation}
When the survival and hazard functions are log-concave (which is the case for exponentially-distributed edge conditional likelihoods, as well as others), then the resulting maximum likelihood inference problem is shown to be convex in~\cite{GomezRodriguez2014}. In fact, the overall maximum likelihood problem decomposes into per-node problems which can be solved using a soft-thresholding algorithm, in a manner similar to~\cite{Meinshausen06}. Furthermore, conditions are provided in~\cite{GomezRodriguez2016} under which the resulting estimate is shown to be consistent (as the number of observed cascades tends to infinity), and sample complexity results are provided, quantifying how quickly the error decays as a function of the number of observed cascades.}

{The above heterogeneous approach requires adopting a parametric model for the edge conditional likelihood, which may be difficult to justify in some settings. The approach described in~\cite{Nan2012} uses kernel methods to estimate the edge conditional likelihoods in a non-parametric manner. More recently, a Bayesian approach to infer a graph topology from diffusion observations has been proposed where the infection time is not directly observed \cite{Shaghaghian16}, but rather the state of each node (susceptible or infected) is a latent variable affecting the statistics of the signal which is observed at each node.}

{In summary, many physically-motivated approaches consider the function $\mathcal{F}(\mathcal{G})$ to be an information propagation model on the network, and generally fall under the bigger umbrella of probabilistic inference of the network of diffusion or epidemic data. Notice, however, that despite its probabilistic nature, such inference is carried out with a specific model of the physical phenomena in mind, instead of using a general probability distribution of the observations considered by statistical models in the previous section.
In addition, for both methods in network tomography and those based on information propagation models, the recovered network typically indicates only the existence of edges and does not promote a specific graph-signal structure. As we shall see, this is a clear difference from the GSP models that are discussed in the following section.}

\section{Graph learning: A signal representation perspective}
\label{sec:gsp}
{There is clearly a growing interest} in the signal processing community to analyze signals that are supported on the vertex set of weighted graphs, leading to the fast-growing field of graph signal processing \cite{Shuman13,Sandryhaila13}. GSP enables the processing and analysis of signals that lie {on structured but irregular domains} by generalizing classical signal processing concepts, tools and methods, such as time-frequency analysis and filtering, on graphs \cite{Shuman13,Sandryhaila13,Ortega18}.

Consider a weighted graph $\mathcal{G}=\{\mathcal{V},\mathcal{E}\}$ with the vertex set $\mathcal{V}$ of cardinality $N$ and edge set $\mathcal{E}$. A graph signal is defined as a function $\mathbf{x}: \mathcal{V} \rightarrow \mathbb{R}^N$ that assigns a scalar value to each vertex. {When the graph is undirected,} the combinatorial or unnormalized graph Laplacian matrix $\mathbf{L}$ is defined as:
\begin{equation}
\mathbf{L}=\mathbf{D}-\mathbf{W},
\label{eq:laplacian}
\end{equation}
where $\mathbf{D}$ is the degree matrix that contains the degrees of the vertices along the diagonal, and $\mathbf{W}$ is the weighted adjacency matrix of $\mathcal{G}$. Since $\mathbf{L}$ is a real and symmetric matrix, it admits a complete set of orthonormal eigenvectors with the associated eigenvalues via the eigencomposition:
\begin{equation}
\mathbf{L} = \boldsymbol{\chi} \mathbf{\Lambda} \boldsymbol{\chi}^T,
\label{eq:eigendecomp}
\end{equation}
where $\boldsymbol{\chi}$ is the eigenvector matrix that contains the eigenvectors as columns, and $\boldsymbol{\Lambda}$ is the eigenvalue matrix $\textbf{diag}(\lambda_0, \lambda_1, \cdots, \lambda_{N-1})$ that contains the eigenvalues along the diagonal. Conventionally, the eigenvalues are sorted in an increasing order, and we have for a connected graph: $0 = \lambda_0 < \lambda_1 \leq \cdots \leq \lambda_{N-1}$.
The Laplacian matrix $\mathbf{L}$ enables a generalization of the notion of frequency and Fourier transform for graph signals \cite{Hammond11}. Alternatively, a graph Fourier transform may also be defined using the adjacency matrix $\mathbf{W}$, and this definition can be used in directed graphs \cite{Sandryhaila13}. Furthermore, both $\mathbf{L}$ and $\mathbf{W}$ can be interpreted as a general class of shift operators on graphs \cite{Sandryhaila13}.

{The above operators are used to represent and process signals on a graph in a similar way as in traditional signal processing.  To see this more clearly, consider two equations of central importance in signal processing: $\mathcal{D}\mathbf{c}=\mathbf{x}$ for the synthesis view and $\mathcal{A}\mathbf{x}=\mathbf{b}$ for the analysis view. In the synthesis view, the signal $\mathbf{x}$ is represented as a linear combination of atoms that are columns of a representation matrix $\mathcal{D}$, with $\mathbf{c}$ being the coefficient vector. 
In the context of GSP, the representation $\mathcal{D}$ of a signal on the graph $\mathcal{G}$ is realized via $\mathcal{F}(\mathcal{G})$, i.e., a function of $\mathcal{G}$.
In the analysis view of GSP, given $\mathcal{G}$ and $\mathbf{x}$ and with a design for $\mathcal{F}$ (that defines $\mathcal{A}$), we study the characteristics of $\mathbf{x}$ encoded in {$\mathbf{b}$}. Examples include the generalization of the Fourier and wavelet transforms for graph signals \cite{Hammond11,Sandryhaila13}, which are defined based on mathematical properties of a given graph $\mathcal{G}$. Alternatively, graph dictionaries can be trained by taking into account information from both $\mathcal{G}$ and $\mathbf{x}$ \cite{Zhang12,Thanou14}. }

{Although most GSP approaches focus on developing techniques for analyzing signals on a predefined or known graph, there is a growing interest in addressing the problem of learning graph topologies from observed signals, especially in the case when the topology is not readily available (i.e., not pre-defined given the application domain). This offers a new perspective to the problem of graph learning, especially by focusing on the representation of the observed signals on the learned graph. 
Indeed, this corresponds to a synthesis view of the signal processing model: given $\mathbf{x}$, with some designs for $\mathcal{F}$ and $\mathbf{c}$, we would like to infer $\mathcal{G}$. Of crucial importance is therefore a model that captures the relationship between the signal representation and the graph, which, together with graph operators such as the adjacency/Laplacian matrices or the graph shift operators \cite{Sandryhaila13}, contributes to specific designs for $\mathcal{F}$. Moreover, assumptions on the structure or properties of $\mathbf{c}$ also play an important role in determining the characteristics of the resulting signal representation. Graph learning frameworks that are developed from a signal representation perspective therefore have the unique advantage of enforcing certain desirable representations of the observed signals, by exploiting the notions of frequency-domain analysis and filtering operations on graphs. }

A graph signal representation perspective is complementary to the existing ones that we discussed in the previous section.  
For instance, from the statistical perspective, the majority of approaches for learning graphical models do not lead directly to a graph topology with non-negative edge weights, {a property that is often desirable in real world applications,} and very little work has studied the case of inferring attractive GMRFs. Furthermore, the joint distribution of the random variables is mostly imposed in a global manner, while it is not easy to encourage localized  behavior (i.e., about a subset of the variables) on the learned graph.
The physics perspective, on the other hand, mostly focuses on a few conventional models such as network diffusion and cascades.
{It remains however an open question how observations that do not necessarily come from a well-defined physical phenomenon can be exploited to infer the underlying structure of the data. The graph signal processing viewpoint introduces one more important ingredient that can be used as a regularizer for complicated inference problems: the frequency or spectral representation of the observations. In what follows, we will review three models for signal representation on graphs, which lead to various methodologies for inferring graph topologies from the observed signals.}

\subsection{Models based on signal smoothness}
\label{sec:smoothness}
\begin{figure}[t]
      \centering
            \subfloat[]	{ \includegraphics[width=5cm]{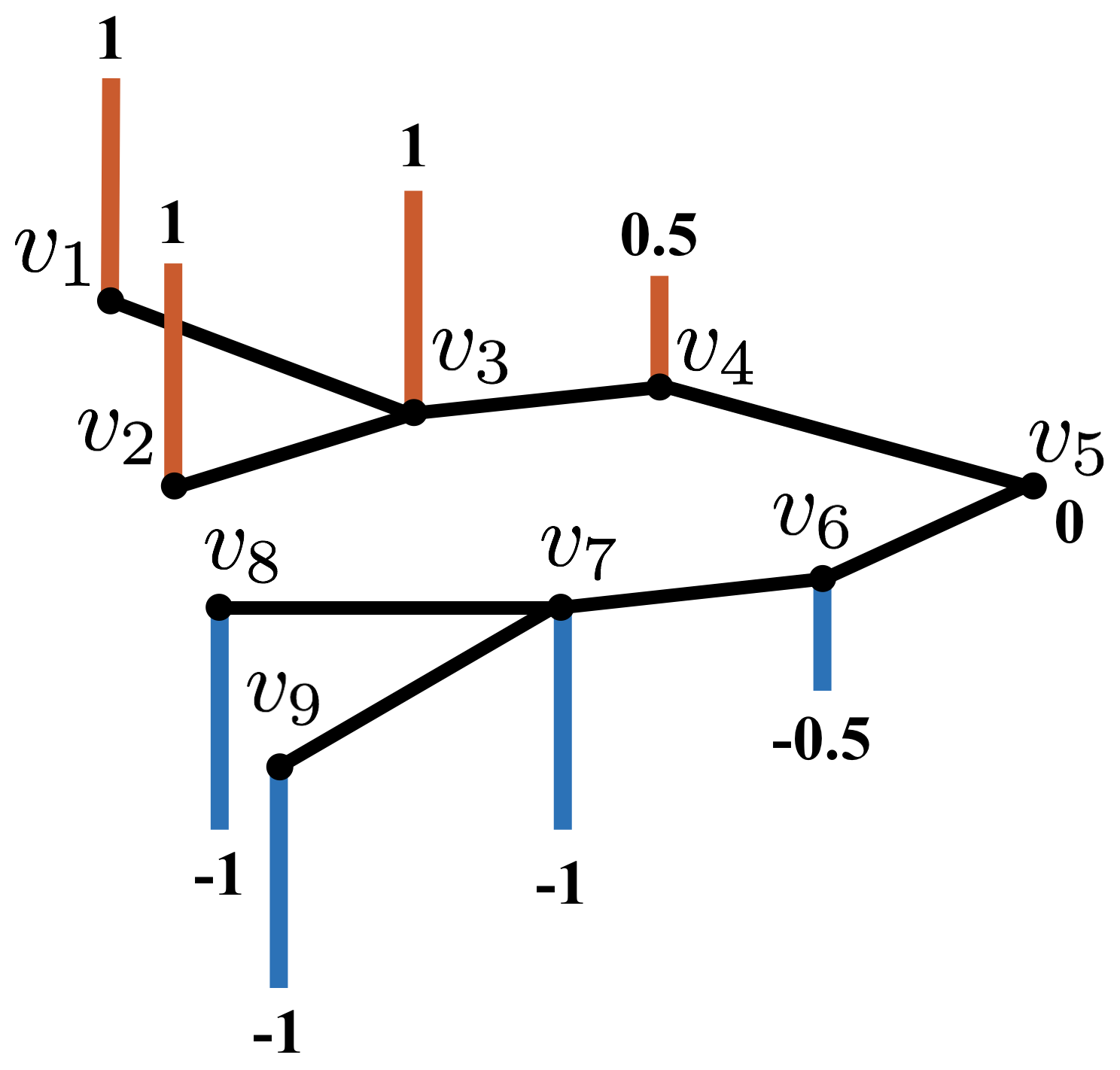}		\label{exa1a}}~~~
            \subfloat[]	{ \includegraphics[width=6cm]{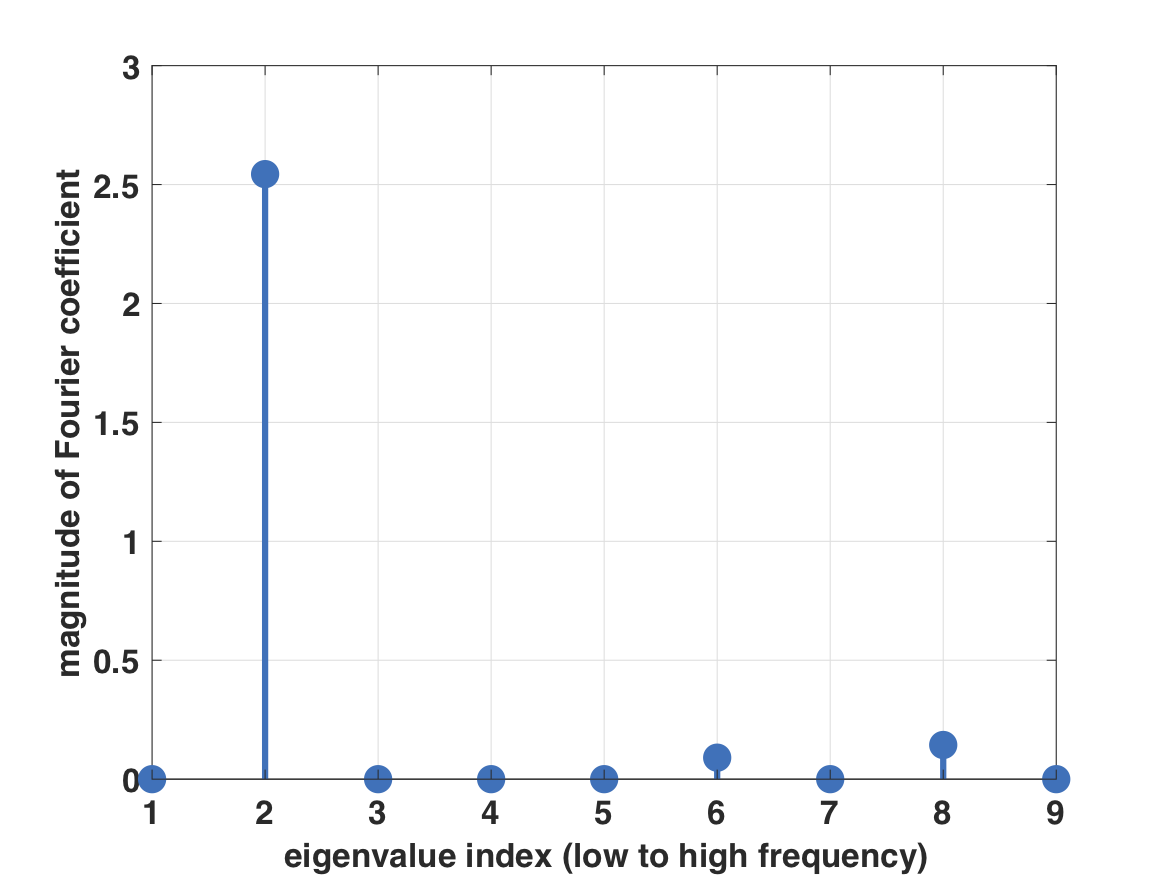}		\label{exa1b}}\\
            \subfloat[]	{ \includegraphics[width=5cm]{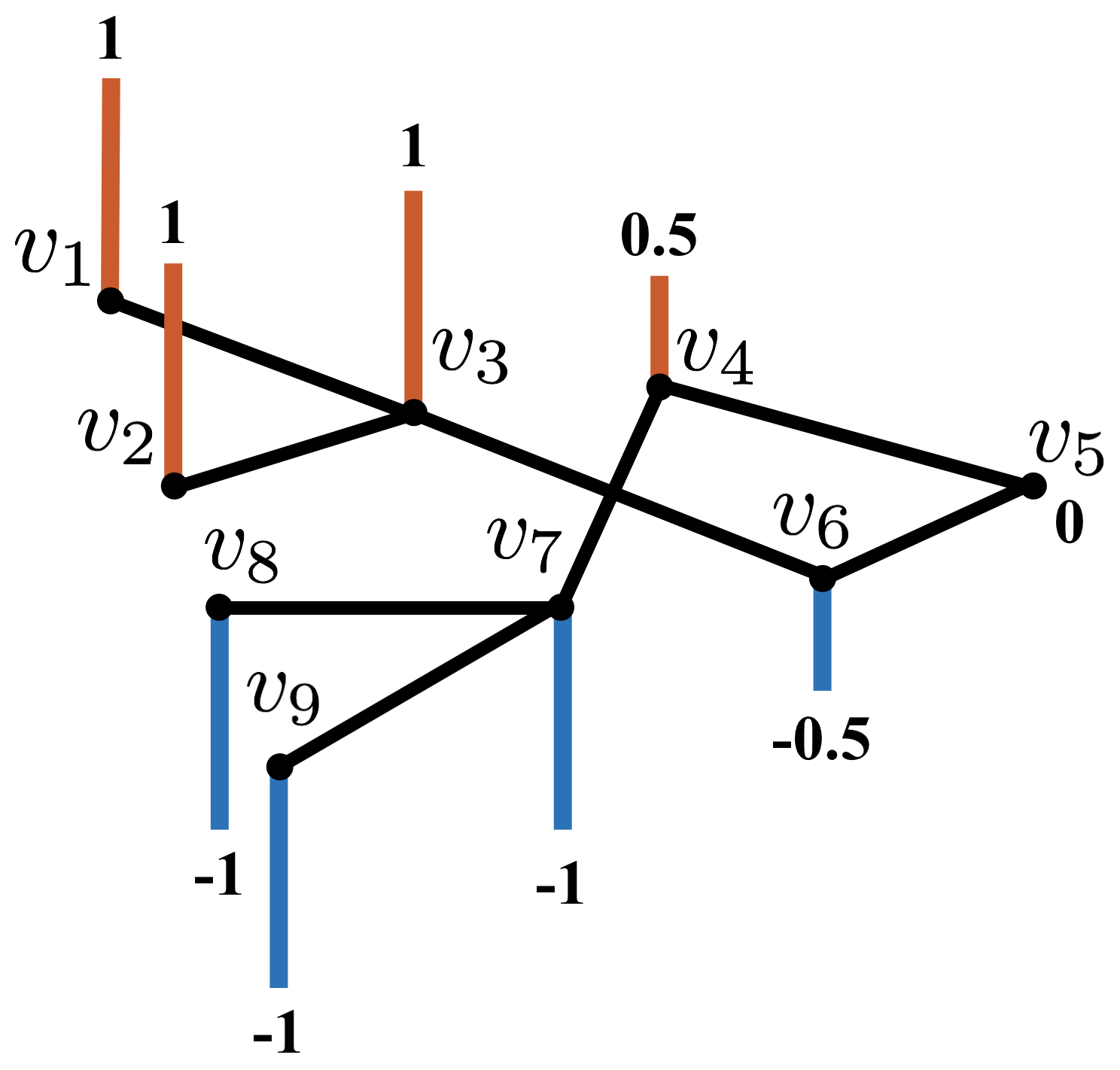}		\label{exa1a}}~~~
            \subfloat[]	{ \includegraphics[width=6cm]{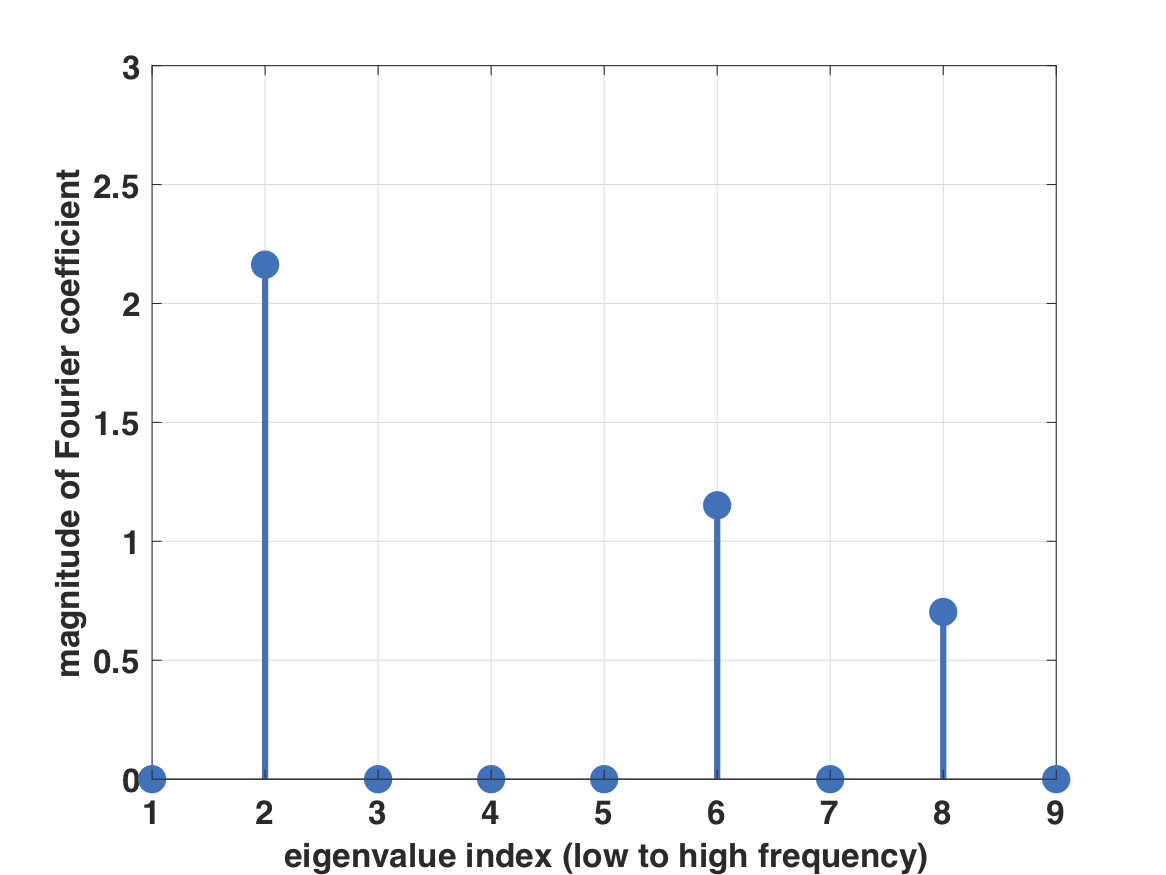}		\label{exa1b}}\\
        \caption{(a) A smooth signal on the graph with $\mathcal{Q}(\mathbf{L})=1$ and (b) its Fourier coefficients in the graph spectral domain. The signal forms a smooth representation on the graph as its values vary slowly along the edges of the graph, and it mainly consists of low frequency components in the graph spectral domain. (c) A less smooth signal on the graph with $\mathcal{Q}(\mathbf{L})=5$ and (d) its Fourier coefficients in the graph spectral domain. A different choice of the graph leads to a different representation of the same signal.}
        \label{fig:exa1}
\end{figure}

The first model we consider is a smoothness model, under which the signal takes similar values at neighboring vertices. Practical examples of this model could be temperature observed at different locations in a flat geographical region, or ratings on movies of individuals in a social network. The measure of smoothness of a signal $\mathbf{x}$ on the graph $\mathcal{G}$ is usually defined by the so-called Laplacian quadratic form:
\begin{equation}
\mathcal{Q}(\mathbf{L}) = \mathbf{x}^T \mathbf{L} \mathbf{x} = \frac{1}{2} \sum_{i,j} w_{ij} \left(\mathbf{x}(i)-\mathbf{x}(j)\right)^2,
\label{eq:lapquad}
\end{equation}
{where $w_{ij}$ is the $ij$-th entry of the adjacency matrix $\mathbf{W}$ and $\mathbf{L}$ is the Laplacian matrix.} Clearly, $\mathcal{Q}(\mathbf{L}) =0$ when $\mathbf{x}$ is a constant signal over the graph (i.e., a DC signal with no variation). More generally, we can see that given the same $L^2$-norm, the smaller the value $\mathcal{Q}(\mathbf{L})$, the more similar are the signal values at neighboring vertices (i.e., the lower the variation of $\mathbf{x}$ is with respect to $\mathcal{G}$). 
One natural criterion is therefore to learn a graph (or equivalently its Laplacian matrix $\mathbf{L}$) such that the signal variation on the resulting graph, i.e., the Laplacian quadratic $\mathcal{Q}(\mathbf{L})$, is small.
As an example, for the same signal, learning a graph in Fig.~\ref{fig:exa1}(a) leads to a smoother signal representation in terms of $\mathcal{Q}(\mathbf{L})$ than that by learning a graph in Fig.~\ref{fig:exa1}(c).
The criterion of minimizing $\mathcal{Q}(\mathbf{L})$ or its variants with powers of $\mathbf{L}$ has been proposed in a number of existing approaches, such as the ones in \cite{Lake10,Daitch09,Hu15}.

{A procedure to} {infer a graph that favors the smoothness of the graph signals can be obtained} using the synthesis model $\mathcal{F}(\mathcal{G}) \mathbf{c}=\mathbf{x}$, and this is the idea behind the approaches in \cite{Dong16,Kalofolias16}. Specifically, consider a factor analysis model with the choice of $\mathcal{F}(\mathcal{G}) = \boldsymbol{\chi}$ and:
\begin{equation}
\mathbf{x} = \boldsymbol{\chi} \mathbf{c} + \boldsymbol{\epsilon},
\end{equation}
where $\boldsymbol{\chi}$ is the eigenvector matrix of the Laplacian $\mathbf{L}$ and $\boldsymbol{\epsilon} \sim \mathcal{N}(\mathbf{0}, \sigma_\epsilon^2 \mathbf{I})$ is additive Gaussian noise. With a further assumption that $\mathbf{c}$ follows a Gaussian distribution with a precision matrix $\mathbf{\Lambda}$: 
\begin{equation}
\mathbf{c} \sim \mathcal{N}(\mathbf{0},\mathbf{\Lambda}^\dagger),
\end{equation}
where $\mathbf{\Lambda}^\dagger$ is the Moore-Penrose pseudo-inverse of the eigenvalue matrix of $\mathbf{L}$, and $\mathbf{c}$ and $\boldsymbol{\epsilon}$ are statistically independent, it is shown in  \cite{Dong16} that the signal $\mathbf{x}$ follows a GMRF model:
\begin{equation}
\mathbf{x} \sim \mathcal{N}(\mathbf{0}, \mathbf{L}^\dagger + \sigma_\epsilon^2 \mathbf{I}).
\end{equation}
This leads to formulating the problem of jointly inferring the graph Laplacian and the latent variable $\mathbf{c}$ as:
\begin{equation}
\min_{\boldsymbol{\chi}, \boldsymbol{\Lambda}, \mathbf{c}} \|\mathbf{x} - \boldsymbol{\chi} \mathbf{c}\|_2^2 + \alpha~\mathbf{c}^T \boldsymbol{\Lambda} \mathbf{c},
\end{equation}
where $\alpha$ is a non-negative regularization parameter related to the assumed noise level $\sigma_\epsilon^2$. By making the change of variables $\mathbf{y} = \boldsymbol{\chi} \mathbf{c}$ and recalling that the matrix of Laplacian eigenvectors $\boldsymbol{\chi}$ is orthornormal, one arrives at the equivalent problem:
\begin{equation}
\min_{\mathbf{L}, \mathbf{y}} \|\mathbf{x} - \mathbf{y} \|_2^2 + \alpha~\mathbf{y}^T \mathbf{L} \mathbf{y},
\end{equation}
in which the Laplacian quadratic form appears.
Therefore, these particular modeling choices for $\mathcal{F}$ and $\mathbf{c}$ lead to a procedure for inferring a graph over which the observation $\mathbf{x}$ is smooth. Note that, there is a one-to-one mapping between the Laplacian matrix $\mathbf{L}$ and a weighted undirected graph, so inferring {$\mathbf{L}$} is equivalent to inferring $\mathcal{G}$.

By taking the matrix form of the observations and adding an $L^2$ penalty, the authors of \cite{Dong16} propose to solve the following optimization problem:
\begin{equation}
\begin{split}
\underset{\mathbf{L},~\mathbf{Y}}{\mbox{minimize}} ~~~ & ||\mathbf{X}-\mathbf{Y}||_F^2 + \alpha~\mathrm{tr}(\mathbf{Y}^T \mathbf{L} \mathbf{Y}) + \beta ||\mathbf{L}||_F^2, \\
\mbox{subject to} ~~~ & \mathrm{tr}(\mathbf{L}) = N,~\mathbf{L} \in \mathcal{L},
\end{split}
\label{eq:smooth}
\end{equation}
where $\mathrm{tr}(\cdot)$ and $||\cdot||_F$ represent the trace and Frobenius norm of a matrix, respectively, and $\alpha$ and $\beta$ are non-negative regularization parameters.
The trace constraint acts as a normalization factor that fixes the volume of the graph and $\mathcal{L}$ is the set of valid Laplacian matrices.
This constitutes the problem of finding $\mathbf{Y}$ that is close to the data observations $\mathbf{X}$, while ensuring at the same time that $\mathbf{Y}$ is smooth on the learned graph represented by its Laplacian matrix $\mathbf{L}$. The Frobenius norm of $\mathbf{L}$ is added to control the distribution of the edge weights and is inspired by the approach in \cite{Hu15}. The problem is solved via alternating minimization in \cite{Dong16}, in which the step of solving for $\mathbf{L}$ bears similarity to the optimization in \cite{Hu15}. A formulation similar to Eq.~(\ref{eq:smooth}) has further been studied in \cite{Kalofolias16} where reformulating the problem in terms of the adjacency matrix $\mathbf{W}$ leads to a more efficient algorithm computationally. Both works emphasize the characteristics of GSP-based graph learning approaches, i.e., enforcing desirable signal representations through the learning process.

As we have seen,
the smoothness property of the graph signal is associated with a multivariate Gaussian distribution, which is also behind the idea of classical approaches for learning graphical models, such as the graphical Lasso. Following the same design for $\mathcal{F}$ and slightly different ones for $\mathbf{\Lambda}$ compared to \cite{Dong16,Kalofolias16}, the authors of \cite{Egilmez17} have proposed to solve a similar objective compared to the graphical Lasso, but with the constraints that the solutions correspond to different types of graph Laplacian matrices (e.g., the combinatorial or generalized Laplacian). {The basic idea in the latter approach is to identify GMRF models such that the precision matrix has the form of a graph Laplacian.
Their work generalizes the classical graphical Lasso formulation and the formulation proposed in \cite{Lake10} to precision matrices restricted to have a Laplacian form. From a probabilistic perspective, the problems of interest correspond to a maximum a posteriori (MAP) parameter estimation of GMRF models, whose precision matrix is a graph Laplacian.  In addition, the proposed approach allows for incorporating prior knowledge on graph connectivity, which, if applicable, can help improve the performance of the graph inference algorithm.}

It is also worth mentioning that, the approaches in \cite{Dong16,Kalofolias16,Egilmez17} learn a graph topology without any explicit constraint on the density of the edges in the learned graph. This information, if available, can be incorporated in the learning process. For example, the work of \cite{Chepuri17} has proposed to learn a graph with a targeted number of edges by selecting the ones that lead to the smallest $\mathcal{Q}(\mathbf{L})$. 

{To summarize, in the global smoothness model, the objective of minimizing the original or a variant of the Laplacian quadratic form of $\mathcal{Q}(\mathbf{L})$ can be interpreted as having $\mathcal{F}(\mathcal{G})=\boldsymbol{\chi}$ and $\mathbf{c}$ following a multivariate Gaussian distribution. However, different learning algorithms may differ in both the output of the algorithm and the computational complexity. For instance, the approaches in \cite{Kalofolias16,Chepuri17} learn an adjacency matrix, while the ones in \cite{Dong16,Egilmez17} learn a graph Laplacian matrix or its variants. In terms of complexity, the approaches in \cite{Dong16}, \cite{Kalofolias16} and \cite{Egilmez17} all solve a quadratic program (QP), with efficient implementations provided in the latter two based on primal-dual techniques and block-coordinate descent algorithms, respectively. On the other hand, the method in \cite{Chepuri17} involves a sorting algorithm that scales with the desired number of edges.}

Finally, it is important to notice that $\mathcal{Q}(\mathbf{L})$ is a measure for \emph{global} smoothness on $\mathcal{G}$ in the sense that a small $\mathcal{Q}(\mathbf{L})$ implies a small variation of signal values along \emph{all} the edges in the graph, and the signal energy is mostly concentrated in the low frequency components in the graph spectral domain. Although global smoothness is often a desirable property for the signal representation, it can also be limiting in other scenarios. The second class of models that we introduce in the following section relaxes this constraint, by allowing for a more flexible representation of the signal in terms of its spectral characteristics.

\subsection{Models based on spectral filtering of graph signals}
\label{sec:filtering}
{The second graph signal model that we consider goes beyond the global smoothness of the signal on the graph and focuses more on the general family of graph signals that are generated by applying a filtering operation to a latent (input) signal. In particular, the filtering operation may correspond to the diffusion of an input signal on the graph.} Depending on the type of the graph filter, and the input signal, the generated signal can have different frequency characteristics (e.g., bandpass signals) and localization properties (e.g., locally smooth signals). Moreover, this family of algorithms is more appropriate {than the one based on a globally smooth signal model} for learning graph topologies when the observations are the result of a diffusion process on a graph.  Particularly, the graph diffusion model can be widely applied in real world scenarios to understand the distribution of heat (sources) \cite{Chung07}, 
such as the propagation of a heat wave in geographical spaces, the movement of people in buildings or vehicles in cities, and the shift of people's interest towards certain subjects on social media platforms \cite{Ma_2008}. 

\begin{figure}[t]
      \centering
        \includegraphics[width=16cm]{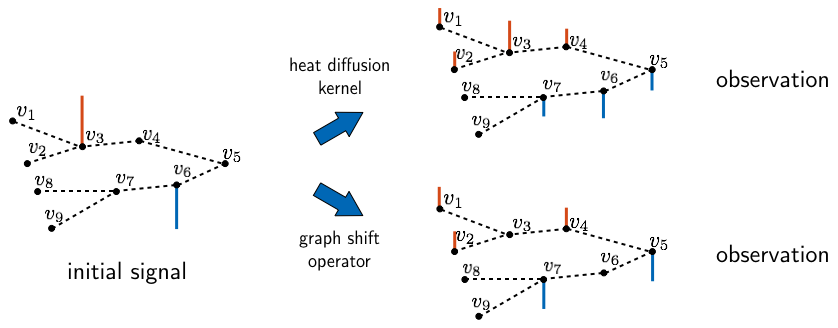}		\label{invcov}
        \caption{Diffusion processes on the graph defined by a heat diffusion kernel (top right) and a graph shift operator (bottom right).}
        \label{fig:diffusion}
\end{figure}

{{In this type of models, the graph filters and the input signals may be interpreted as the functions $\mathcal{F}(\mathcal{G})$ and the coefficients $\mathbf{c}$ in our synthesis model, respectively.
The existing methods in the literature therefore differ in the assumptions on $\mathcal{F}$ as well as the distribution of $\mathbf{c}$.} In particular, $\mathcal{F}$ may be defined as an arbitrary (polynomial) function of a matrix related to the graph \cite{Segarra17a,Pasdeloup18}, or a well-known diffusion kernel such as the heat diffusion kernel \cite{Thanou17} (see Fig.~\ref{fig:diffusion} for two examples). The assumptions on $\mathbf{c}$ can also vary, with the most prevalent ones being zero-mean Gaussian distribution, and sparsity.  Broadly speaking, we can distinguish the graph learning algorithms belonging to this family in two different categories. {The first category models the graph signals as stationary processes on graphs, where the eigenvectors of a graph operator, such as the adjacency/Laplacian matrix or a shift operator, are estimated from the sample covariance matrix of the observations in the first step. The eigenvalues are then estimated in the second step to obtain the operator.}
The second category poses the graph learning problem as a dictionary learning problem with a prior on the coefficients $\mathbf{c}$. In what follows, we will give a few representative examples of both categories, which differ in terms of graph filters as well as input signal characteristics.}

\subsubsection{{Stationarity based learning frameworks}} The main characteristic of this line of work is that, given a stationarity assumption, the eigenvectors of a graph operator are estimated by the empirical covariance matrix of the observations. In particular, the graph signal $\mathbf{x}$ can be generated from:   
\begin{equation}
\label{eq:diff_model} \mathbf{x} = \beta_0 \Pi_{k = 1}^{\infty} (\mathbf{I} - \beta_k \mathbf{S})\mathbf{c} = \sum_{k = 0}^{\infty}\alpha_k \mathbf{S}^k \mathbf{c},
\end{equation}
for some set of the parameters $\{\alpha\}$ and $\{\beta\}$. The latter implies that there exists an underlying diffusion process in the graph operator $\mathbf{S}$, which can be the adjacency matrix, Laplacian, or a variation thereof, that produces the signal $\mathbf{x}$ from the input signal $\mathbf{c}$.  By assuming a finite polynomial degree $K$, the generative signal model becomes:
\begin{equation}
\mathbf{x} = \mathcal{F}(\mathcal{G})\mathbf{c} = \sum_{k = 0}^{K}\alpha_k \mathbf{S}^k \mathbf{c},
\end{equation}
where the connectivity matrix of $\mathcal{G}$ is captured through the graph operator $\mathbf{S}$.
Usually, $\mathbf{c}$ is assumed to be a zero-mean graph signal with covariance matrix $\mathbf{\Sigma}_c = \mathbb{E}[\mathbf{c} \mathbf{c}^T]$. In addition, if $\mathbf{c}$ is white and $\boldsymbol{\Sigma}_c = \mathbf{I}$, Eq.~(\ref{eq:diff_model}) is equivalent to assuming that the graph process $\mathbf{x}$ is \emph{stationary} in $\mathbf{S}$. This assumption of stationarity is important for estimating the eigenvectors of the graph operator. Indeed, since the graph operator $\mathbf{S}$ is often a real and symmetric matrix, its eigenvectors are also eigenvectors of the covariance matrix $\boldsymbol{\Sigma}_x$. As a matter of fact:
\begin{equation}
\begin{split}
\boldsymbol{\Sigma}_x &= \mathbb{E}[\mathbf{x} \mathbf{x}^T] = \mathbb{E}\left[\sum_{k = 0}^{K}\alpha_k \mathbf{S}^k \mathbf{c} \big(\sum_{k = 0}^{K}\alpha_k \mathbf{S}^k \mathbf{c}\big)^T\right]\\
 &= \sum_{k = 0}^{K}\alpha_k \mathbf{S}^k \big(\sum_{k = 0}^{K}\alpha_k \mathbf{S}^k\big)^T  = \boldsymbol{\chi} \left(\sum_{k = 0}^{K}\alpha_k \mathbf{\Lambda}^k\right)^2 \boldsymbol{\chi}^T, 
\end{split} 
\label{eq:stationarity_cov_step2}
\end{equation}
where we have used the assumption that $\mathbf{\Sigma}_c = \mathbf{I}$ and the eigendecomposition $\mathbf{S} = \boldsymbol{\chi} \mathbf{\Lambda} \boldsymbol{\chi}^T$. Given a sufficient number of graph signals, the eigenvectors of the graph operator $\mathbf{S}$ can therefore be approximated by the eigenvectors of the empirical covariance matrix of the observations. {To recover $\mathbf{S}$, the second step of the process would then be to learn its eigenvalues.}

The authors in \cite{Pasdeloup18} follow the aforementioned reasoning and model the diffusion process by powers of the normalized Laplacian matrix.  More precisely, they propose an algorithm for characterizing and then computing a set of admissible diffusion matrices, which defines a polytope. In general, this polytope corresponds to a continuum of graphs that are all consistent with the observations. To obtain a particular solution, an additional criterion is required. Two such criteria are proposed: one which encourages the resulting graph to be sparse, and another which encourages the recovered graph to be \emph{simple} (i.e., a graph in which no vertex has a connection to itself hence an adjacency matrix with only zeros along the diagonal).

Similarly, in \cite{Segarra17a}, after obtaining the eigenvectors of a graph shift operator, the graph learning problem is equivalent to learning its eigenvalues, under the constraints that the shift operator obeys some desired properties such as sparsity.
The optimization problem of \cite{Segarra17a} can be written as: 
\begin{equation}
\begin{split}
\underset{\mathbf{S} ,~\mathbf{\Psi}}{\mbox{minimize}} ~~~ & f(\mathbf{S}, \mathbf{\Psi}), \\
\mbox{subject to} ~~~ & \mathbf{S} = \mathbf{\chi} \mathbf{\Psi} \mathbf{\chi}^T, ~\mathbf{S} \in \mathcal{S},
\end{split}
\label{eq:opt_prob_Segarra}
\end{equation}
where $f(\cdot)$ is a convex function applied on $\mathbf{S}$ that imposes the desired properties of $\mathbf{S}$, e.g., sparsity via an {entry-wise} $L^1$-norm, and $\mathcal{S}$ is the constraint set of $\mathbf{S}$ being a valid graph operator, e.g., non-negativity of the edge weights.
The stationarity assumption is further relaxed in \cite{Shafipour18a}. However, all these approaches are based on the assumption that the sample covariance of the observed data and the graph operator have the same set of eigenvectors. Thus, their performance depends on the accuracy of eigenvectors obtained from the sample covariance of data, which can be difficult to guarantee especially when the number of data samples is small relative to the number of vertices in the graph.

Given the limitation in estimating the eigenvectors of the graph operator from the sample covariance, the work of \cite{Egilmez18} has proposed a different approach. 
They have formulated the problem of graph learning as a graph system identification problem where, by assuming that the observed signals are output of a system with a graph-based filter given certain input,
the goal is to learn a weighted graph (a graph Laplacian matrix) and the graph-based filter (a function of the graph Laplacian matrices). {The algorithm is based on the minimization of a regularized maximum likelihood criterion and it is valid under the assumption that the graph filters are one-to-one functions, i.e., increasing or decreasing in the space of eigenvalues, such as a heat diffusion kernel. 
More specifically, the system input is assumed to be multivariate white Gaussian noise (hence the stationarity assumption on the observed signals), and Eq.~(\ref{eq:stationarity_cov_step2}) is again used for computing an initial estimate of the eigenvectors. However, different from \cite{Segarra17a,Pasdeloup18} where these eigenvectors are used directly in forming the graph operators, in \cite{Egilmez18} they are used to compute the graph Laplacian:
after initializing the filter parameter, the algorithm iterates until convergence between the following three steps:
(a) pre-filter the sample covariance using the inverse of the graph filter; (b) estimate a graph Laplacian from the pre-filtered covariance matrix by solving a maximum likelihood optimization criterion, using an algorithm proposed in \cite{Egilmez17}; (c) update the filter parameter based on the current estimate of the graph Laplacian.
Compared to \cite{Segarra17a,Pasdeloup18}, this approach may therefore lead to a more accurate inference of the graph operator (graph Laplacian in this case).

\subsubsection{{Graph dictionary based learning frameworks}}
{Methods belonging to this category are based on the notion of spectral graph  dictionaries for efficient signal representation.
Specifically, the authors in \cite{Thanou17,Maretic17} assume a different graph signal diffusion model,} where the data consist of (sparse) combinations of overlapping local patterns that reside on the graph. These patterns may describe localized events or specific processes appearing at different vertices of the graph, such as traffic bottlenecks in transportation networks or rumor sources in social networks. The graph signals are then viewed as  observations at different time instants of a few processes that start at different nodes of an unknown graph and diffuse with time. Such signals can be represented as the combination of graph heat kernels or, more generally, of localized graph kernels. {Both algorithms can be considered as a generalization of dictionary learning to graph signals. Dictionary learning \cite{Rubinstein10,Tosic11} is an area of research in signal processing and machine learning where the signals are represented as a linear combination of simple components, i.e., atoms, in an (often) overcomplete basis. Signal decompositions with overcomplete dictionaries offer a way to efficiently approximate or process signals, such that the  important characteristics  are revealed by the sparse signal representation. Due to these desirable properties, dictionary learning has been extended to the representation of graph signals, and eventually has been applied to the problem of graph inference.}

{Next, we provide more details on one of the above mentioned algorithms.} The authors in \cite{Thanou17} have focused on graph signals generated from heat diffusion processes, which are useful in identifying processes evolving nearby a starting seed node. An illustrative example of such a signal can be found in Fig.~\ref{fig:diffusion_example}, in which case the graph Laplacian matrix is used to model the diffusion of the heat throughout a graph.
The concatenation of a set of heat diffusion operators at different time instances defines a graph dictionary that is further on used to represent the graph signals.
Hence, the graph signal model becomes:
\begin{equation}
\mathbf{x} = \mathcal{F}(\mathcal{G}) \mathbf{c} = [e^{-\tau_1 \mathbf{L}} ~ e^{-\tau_2 \mathbf{L}} ~ \cdots ~ e^{-\tau_S \mathbf{L}} ~ ]\mathbf{c} = \sum_{s=1}^S e^{-\tau_s \mathbf{L}}\mathbf{c}_s,
\end{equation} 
which is a linear combination of different heat diffusion processes evolving on the graph. In this synthesis model, the coefficients $\mathbf{c}_s$ corresponding to a subdictionary $e^{-\tau_s \mathbf{L}}$ can be seen as a graph signal that goes through a heat diffusion process on the graph. The signal component $e^{-\tau_s \mathbf{L}} \mathbf{c}_s$  can then be interpreted as the result of this diffusion process at time $\tau_s$.
It is interesting to notice that the parameter $\tau_s$ in the model carries a notion of scale. In particular, when $\tau_s $ is small, the $i$-th column of $e^{-\tau_s \mathbf{L}}$, i.e., the atom centered at node $v_i$ of the graph, is mainly localized in a small neighborhood of $v_i$. As $\tau_s$ becomes larger, it reflects information about the graph at a larger scale around $v_i$. Thus, the signal model can be seen as an additive model of diffusion processes of $S$ initial graph signals, that undergo a diffusion model with different diffusion times.

\begin{figure*}[!t]
      \centering
      {\includegraphics[width=16cm]{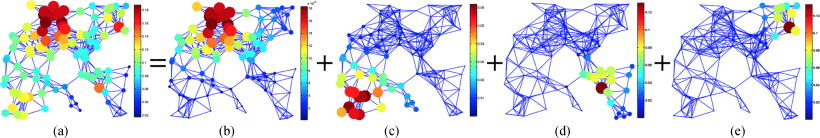}	\label{heatdiff}}
      \caption{(a) A graph signal. (b-e) Its decomposition in four localized simple components. Each component is a heat diffusion process $(e^{-\tau\mathbf{L}})$ at time $\tau$ that has started from different network nodes. The size and the color of each ball indicate the value of the signal at each vertex of the graph. Figure from \cite{Thanou17}.}
        \label{fig:diffusion_example}
\end{figure*}

An additional assumption on the above signal model is that the diffusion processes are expected to start from only a few nodes of the graph, at specific times, and spread over the entire graph over time\footnote{When no locality assumptions are imposed (e.g., large $\tau_s$) and a single diffusion kernel is used in the dictionary, the model reduces to a global smoothness model.}.
This assumption can be formally captured by imposing a sparsity constraint on the latent variable $\mathbf{c}$.
The graph learning problem can be cast as a structured dictionary learning problem, where the dictionary is defined by the unknown graph Laplacian matrix. The latter can then be estimated as a solution of the following optimization problem:
\begin{equation}
\begin{split}
\underset{\mathbf{L} ,~\mathbf{C}, ~\tau}{\mbox{minimize}} ~~~ & \| \mathbf{X} -\D \mathbf{C} \|^{2}_{F} +  \alpha \sum_{m = 1}^M\|\mathbf{c}_m\|_1 + \beta \|\mathbf{L}\|_F^2, \\
\mbox{subject to} ~~~ & \D = [e^{-\tau_1 \mathbf{L}} ~ e^{-\tau_2 \mathbf{L}} \dots e^{-\tau_S \mathbf{L}} ~ ], ~ \{\tau_s\}_{s=1}^S \ge 0, \\
~~~ & \mathrm{tr}(\mathbf{L}) = N, ~ \mathbf{L} \in \mathcal{L},
\end{split}
\label{eq:opt_prob_heat_kernel}
\end{equation}
where the constraints on $\mathcal{L}$ is the same as that in Eq.~(\ref{eq:smooth}).
Following the same reasoning, the work in \cite{Maretic17} extends the heat diffusion dictionary to the more general family of polynomial graph kernels. In summary, these approaches propose to recover the graph Laplacian matrix by assuming that the graph signals can be sparsely represented by a dictionary that consists of graph diffusion kernels.

{In summary, from the perspective of spectral filtering, and in particular network diffusion, the function $\mathcal{F}(\mathcal{G})$ is one that helps define a meaningful diffusion process on the graph via the graph Laplacian, heat diffusion kernel, or other more general graph shift operators. This directly leads to the slightly different output of the learning algorithms in \cite{Pasdeloup18,Segarra17a,Thanou17}.
The choice of the coefficients $\mathbf{c}$, on the other hand, determines specific characteristics of the graph signals, such as stationarity or sparsity.
In terms of computational complexity, the methods in \cite{Pasdeloup18,Segarra17a,Thanou17} all involve the computation of eigenvectors, followed by solving a linear program (LP), a semidefinite program (SDP), and a SDP, respectively.}

\subsection{Models based on causal dependencies on graphs}
\label{sec:causal}
The models described in the previous two sections are mainly designed for learning undirected graphs, which is also the predominant consideration in the current GSP literature. 
Undirected graphs are associated with symmetric Laplacian matrices $\mathbf{L}$, which admit a complete set of orthonormal eigenvalues and eigenvectors that conveniently provide a notion of frequency for signals on graphs.
It is often the case, however, that in some application domains learning directed graphs is more desirable as in those cases the directions of edges may be interpreted as causal dependencies between the variables that the vertices represent. For example, in brain analysis, even though the inference of an undirected \emph{functional connectivity} between the regions of interest (ROIs) is certainly of interest, a directed \emph{effective connectivity} may reveal extra information about the causal dependencies between those regions \cite{Friston94,Shen16}. The third class of models that we discuss is therefore one that allows for the inference of such directed dependencies.

The authors of \cite{Mei17} have proposed a causal graph process based on the idea of sparse vector autoregressive (SVAR) estimation \cite{Songsiri10,Bolstad11}. In their model, the signal at time step $t$, $\mathbf{x}[t]$, is represented as a linear combination of its observations in the past $T$ time steps and a random noise process $\mathbf{n}[t]$:
\begin{equation}
\begin{split}
\mathbf{x}[t] &= \mathbf{n}[t] + \sum_{j=1}^T P_j(\mathbf{W}) \mathbf{x}[t-j]\\
&= \mathbf{n}[t] + \sum_{j=1}^T \sum_{k=0}^j a_{jk} {\mathbf{W}}^k \mathbf{x}[t-j],
\end{split}
\label{eq:svar-model}
\end{equation}
\noindent where $P_j(\mathbf{W})$ is a degree $j$ polynomial of the (possibly directed) adjacency matrix $\mathbf{W}$ with coefficients $a_{jk}$ (see Fig.~\ref{fig:temporal} for an illustration).
Clearly, this model admits the design of $\mathcal{F}(\mathcal{G})=P_i(\mathbf{W})$ and $\mathbf{c}=\mathbf{x}[t-i]$ in forming one time-lagged copy of the signal $\mathbf{x}[t]$. 
For temporal observations $\mathbf{X}= \big( \mathbf{x}[0]~\mathbf{x}[1]~\cdots~\mathbf{x}[M-1] \big)$, the authors have therefore proposed to solve the following optimization problem:
\begin{equation}
\min_{\mathbf{W},\mathbf{a}} ~ \frac{1}{2} \sum_{t=T}^{M-1} \Big\| \mathbf{x}[t] - \sum_{j=1}^T P_j(\mathbf{W}) \mathbf{x}[t-j] \Big\|_2^2 + \alpha~||\text{vec}(\mathbf{W})||_1 + \beta~||\mathbf{a}||_1,
\label{eq:svar}
\end{equation}
where $\text{vec}(\mathbf{W})$ is the vectorized form of $\mathbf{W}$, $\mathbf{a} = \big( a_{10}~a_{11}~\cdots~a_{jk}~\cdots~a_{TT} \big)$ is a vector of all the polynomial coefficients $a_{jk}$, and the {entry-wise} $L^1$-norm is imposed on $\mathbf{W}$ and $\mathbf{a}$ for promoting sparsity. Due to non-convexity introduced by the matrix polynomials, the problem in Eq.~(\ref{eq:svar}) is solved in three steps, i.e., solving sequentially for $P_j(\mathbf{W})$, $\mathbf{W}$, and $\mathbf{a}$. In summary, in the SVAR model, the specific designs of $\mathcal{F}$ and $\mathbf{c}$ lead to a particular generative process of the observed signals on the learned graph. Similar ideas can also be found in the Granger causality or vector autoregressive models (VARMs) \cite{Roebroeck05,Goebel03}.

\begin{figure}[t]
      \centering
        \includegraphics[width=16cm]{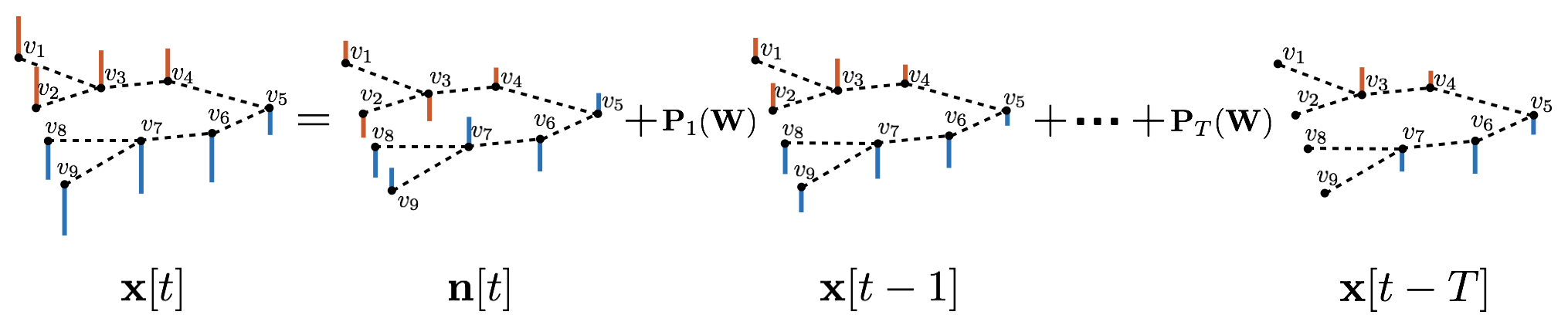}		\label{temporal}
        \caption{A graph signal $\mathbf{x}$ at time step $t$ is modeled as a linear combination of its observations in the past $T$ time steps and a random noise process $\mathbf{n}[t]$.}
        \label{fig:temporal}
\end{figure}

Structural equation models (SEMs) are another popular approach for inferring directed graphs \cite{Kaplan09,Mclntosh94}. In the SEMs, the signal observation $\mathbf{x}$ at time step $t$ is modeled as:
\begin{equation}
\mathbf{x}[t] = \mathbf{W} \mathbf{x}[t] + \mathbf{E} \mathbf{y}[t] + \mathbf{n}[t],
\label{eq:sem}
\end{equation}
where the first term in Eq.~(\ref{eq:sem}) consists of endogenous variables, which define the signal value at each variable as a linear combination of the values at its neighbors in the graph, and the second term represents exogenous variables $\mathbf{y}[t]$ with a coefficient matrix $\mathbf{E}$. The third term represents observation noise which is similar to that in Eq.~(\ref{eq:svar-model}). The endogenous component of the signal implies a choice of $\mathcal{F}(\mathcal{G}) = \mathbf{W}$ (which can again be directed) and $\mathbf{c} = \mathbf{x}[t]$ and, similar to the SVAR model, enforces a certain generative process of the signal on the learned graph. 

{As we can see, causal dependencies on the graph, either between different components of the signal or between its present and past observations, can be conveniently modeled in a straightforward manner by choosing $\mathcal{F}(\mathcal{G})$ as a polynomial of the adjacency matrix of a directed graph and choosing the coefficients $\mathbf{c}$ as the present or past signal observations.
As a consequence, methods in \cite{Mei17,Baingana17,Shen16} are all able to learn an asymmetric graph adjacency matrix, which is a potential advantage compared to methods based on the previous two models.
Furthermore, the SEMs can be extended to track network topologies that evolve dynamically \cite{Baingana17} and deal with highly correlated data \cite{Traganitis17}, or combined with the SVAR model which leads to the structural vector autoregressive models (SVARMs) \cite{Chen11}. 
Interested readers are referred to \cite{Giannakis18} for a recent review of the related models.
In these extensions of the classical models, the designs of $\mathcal{F}$ and $\mathbf{c}$ can be generalized accordingly to link the signal representation and the learned graph topology.
Finally, as an overall comparison, the differences between methods that are based on the three models discussed in this review are summarized in Table~\ref{tab:comparison}.}

\begin{table}[t]
\centering
\caption{Comparison between different GSP-based approaches to graph learning.}
\label{tab:comparison}
\scalebox{0.8}{
\begin{tabular}{|c|c|c|c|c|c|}
\hline
\multirow{2}{*}{\textbf{Method}} & \multirow{2}{*}{\textbf{Signal Model}}                                                            & \multicolumn{2}{c|}{\textbf{Assumption}}                                                                                                         & \multirow{2}{*}{\textbf{Learning Output}}                                       & \multirow{2}{*}{\textbf{Edge Directionality}} \\ \cline{3-4}
                                  &                                                                                                   & $\mathcal{F}(\mathcal{G})$                                                                 & $\textbf{c}$                                                          &                                                                                 &                                          \\ \hline
Dong et al. \cite{Dong16}                & Global Smoothness                                                                                 & \begin{tabular}[c]{@{}c@{}}Eigenvector \\ Matrix\end{tabular}              & Gaussian                                                            & Laplacian                                                                       & Undirected                               \\ \hline
Kalofolias et al. \cite{Kalofolias16}          & Global Smoothness                                                                                 & \begin{tabular}[c]{@{}c@{}}Eigenvector \\ Matrix\end{tabular}              & Gaussian                                                            & Adjacency Matrix                                                                & Undirected                               \\ \hline
Egilmez et al. \cite{Egilmez17}             & Global Smoothness                                                                                 & \begin{tabular}[c]{@{}c@{}}Eigenvector \\ Matrix\end{tabular}              & Gaussian                                                            & Generalized Laplacian                                                           & Undirected                               \\ \hline
Chepuri et al. \cite{Chepuri17}             & Global Smoothness                                                                                 & \begin{tabular}[c]{@{}c@{}}Eigenvector \\ Matrix\end{tabular}              & Gaussian                                                            & Adjacency Matrix                                                                & Undirected                               \\ \hline
Pasdeloup et al. \cite{Pasdeloup18}           & \begin{tabular}[c]{@{}c@{}}Spectral Filtering \\ (Diffusion by Adjacency)\end{tabular}            & \begin{tabular}[c]{@{}c@{}}Normalized \\ Adjacency Matrix\end{tabular}     & IID Gaussian                                                     & \begin{tabular}[c]{@{}c@{}}Normalised Adjacency Matrix\\ Normalized Laplacian\end{tabular} & Undirected                               \\ \hline
Segarra et al. \cite{Segarra17a}             & \begin{tabular}[c]{@{}c@{}}Spectral Filtering \\ (Diffusion by Graph Shift Operator)\end{tabular} & \begin{tabular}[c]{@{}c@{}}Graph Shift \\ Operator\end{tabular}            & IID Gaussian                                                     & Graph Shift Operator                                                            & Undirected                               \\ \hline
Thanou et al. \cite{Thanou17}              & \begin{tabular}[c]{@{}c@{}}Spectral Filtering \\ (Heat diffusion)\end{tabular}                    & Heat Kernel                                                                & Sparsity                                                            & Laplacian                                                                       & Undirected                               \\ \hline
Mei and Moura \cite{Mei17}              & \begin{tabular}[c]{@{}c@{}}Causal Dependency \\ (SVAR)\end{tabular}                               & \begin{tabular}[c]{@{}c@{}}Polynomials of \\ Adjacency Matrix\end{tabular} & Past Signals                                                        & Adjacency Matrix                                                                & Directed                                 \\ \hline
Baingana et al. \cite{Baingana17}            & \begin{tabular}[c]{@{}c@{}}Causal Dependency  \\ (SEM)\end{tabular}                               & Adjacency Matrix                                                           & Present Signal                                                      & \begin{tabular}[c]{@{}c@{}}Time-Varying \\ Adjacency Matrix\end{tabular}        & Directed                                 \\ \hline
Shen et al. \cite{Shen16}                & \begin{tabular}[c]{@{}c@{}}Causal Dependency \\ (SVARM)\end{tabular}                              & \begin{tabular}[c]{@{}c@{}}Polynomials of \\ Adjacency Matrix\end{tabular} & \begin{tabular}[c]{@{}c@{}}Past and \\ Present Signals\end{tabular} & Adjacency Matrix                                                                & Directed                                 \\ \hline
\end{tabular}}
\end{table}

\subsection{Connections with the broader literature}
We have seen that GSP-based approaches can be unified by the viewpoint of learning graph topologies that enforce desirable representations of the signals on the learned graph. This offers a new interpretation of the traditional statistical and physically-motivated models. First, as a typical example of approaches for learning graphical models, the graphical Lasso solves the optimization problem of Eq.~(\ref{eq:gLasso}) in which the trace term $\mathrm{tr}(\widehat{\boldsymbol{\Sigma}} \mathbf{\Theta}) = \frac{1}{M-1}\mathrm{tr} (\mathbf{X}^T \mathbf{\Theta} \mathbf{X})$ bears similarity to the Laplacian quadratic form $\mathcal{Q}(\mathbf{L})$ and the trace term in the problem of Eq.~(\ref{eq:smooth}), when the precision matrix $\mathbf{\Theta}$ is chosen to be the graph Laplacian $\mathbf{L}$. This is the case for the approach in \cite{Lake10}, which has proposed to consider $\mathbf{\Theta} = \mathbf{L} + \frac{1}{\sigma^2}\mathbf{I}$ (see Eq.~(\ref{eq:lake})) as a regularized Laplacian to fit into the formulation of Eq.~(\ref{eq:gLasso}). The graphical Lasso approach therefore can be interpreted as one that promotes global smoothness of the signals on the learned topology.

Second, models based on spectral filtering and causal dependencies on graphs can generally be thought of as the ones that define generative processes of the observed signals, in particular the diffusion processes on the graph.
This is achieved either explicitly by choosing $\mathcal{F}(\mathcal{G})$ as diffusion matrices as that in Section~\ref{sec:filtering}, or implicitly by defining the causal processes of signal generation as that in Section~\ref{sec:causal}. Both types of models share similar philosophies as the ones developed from a physics viewpoint in Section~\ref{sec:physics}, in that they all propose to infer the graph topologies by modeling signals as outcomes of physical processes on the graph, especially the diffusion and cascading processes.

It is also interesting to notice that certain models can be interpreted from all the three viewpoints, an example being the global smoothness model. Indeed, in addition to the statistical and GSP perspectives described above, the property of global smoothness can also be observed in a square-lattice Ising model \cite{Cipra87}, hence admitting a physical interpretation. 
{Despite the connections with traditional approaches, however,} GSP-based approaches offer some unique advantages compared to the classical methods. On the one hand, the flexibility in designing the function $\mathcal{F}(\mathcal{G})$ allows for statistical properties of the observed signals that are not limited to a Gaussian distribution, which is however the predominant choice in many statistical machine learning methods. On the other hand, this also makes it easier to consider models that go beyond a simple diffusion or cascade model. For example, by the sparsity assumption on the coefficients $\mathbf{c}$, the method in \cite{Thanou17} defines the signals as the outcomes of possibly more than one diffusion processes originated at different parts of the graph after possibly different time steps. 
Similarly, by choosing different $\mathcal{F}(\mathcal{G})$ and $\mathbf{c}$, the SVAR models \cite{Mei17} and the SEMs \cite{Baingana17} correspond to different generative processes of the signals, one based on the static network structure and the other on temporal dynamics. These design flexibilities provide more powerful modeling of the signal representation for the graph inference process.

\section{Applications of GSP-based graph learning methods}
The field of GSP is strongly motivated by a wide range of applications where there exist inherent structures behind data observations. Naturally, GSP-based graph learning methods are appealing in areas where learning hidden structures behind data has been of constant interest. 
In particular, the emphasis on the modeling of the signal representation within the learning process has made them increasingly popular in a growing number of applications. 
Currently, these methods mainly find applications in image coding and compression, brain signal analysis, and a few other diverse areas, {as described briefly below.}

\subsection{Image coding and compression}
Image representation and coding has been one main area of interest for GSP-based methods. Images can be naturally thought of as graph signals defined on a regular grid structure, where the nodes are the image pixels and the edge weights capture the similarity between adjacent pixels. The design of new flexible graph signal representations has opened the door to new structure-aware transform coding techniques, and eventually to more efficient image compression frameworks \cite{Cheung18}. Such representation permits to go beyond traditional transform coding by moving from classical fixed transforms such as the discrete cosine transform (DCT) to graph-based transforms that are better adapted to the actual image structure.

The design of the graph and the corresponding transform remains, however, one of the biggest challenges in graph-based image compression. A suitable graph for effective transform coding should lead to easily compressible signal coefficients, at the cost of a small overhead for coding the graph. 
Most graph-based coding techniques focus mainly on images, and they construct the graph by considering pairwise similarities among pixel intensities.  A few attempts on adapting the graph topology and consequently the graph transform exist in the literature, as for example in \cite{Hu15Compression,Rotondo15}. However, they rely on the selection from a set of  representative graph templates, without being fully adapted to the image signals. 

\begin{figure}[t]
      \centering
      \subfloat[]
        {\includegraphics[width=12.6cm]{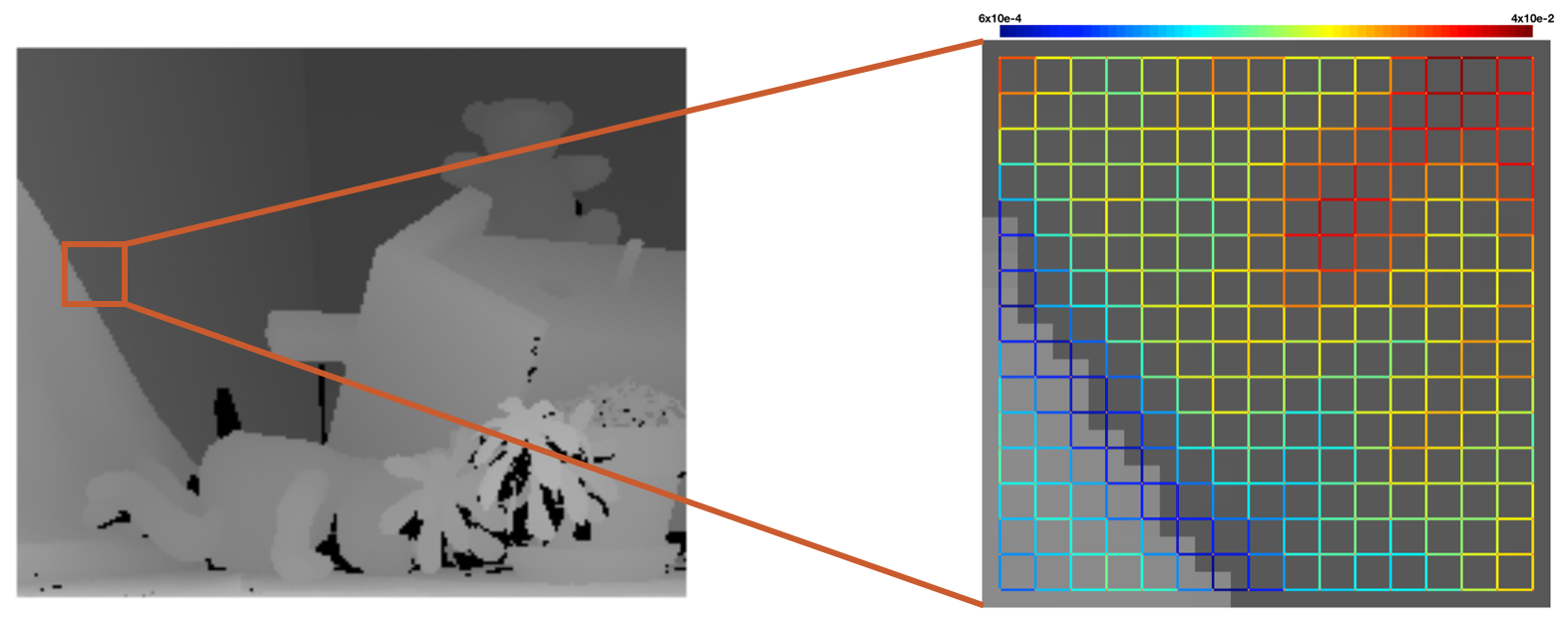}		\label{compression-example-image}}\\
      \subfloat[]
		{ \includegraphics[width=7cm]{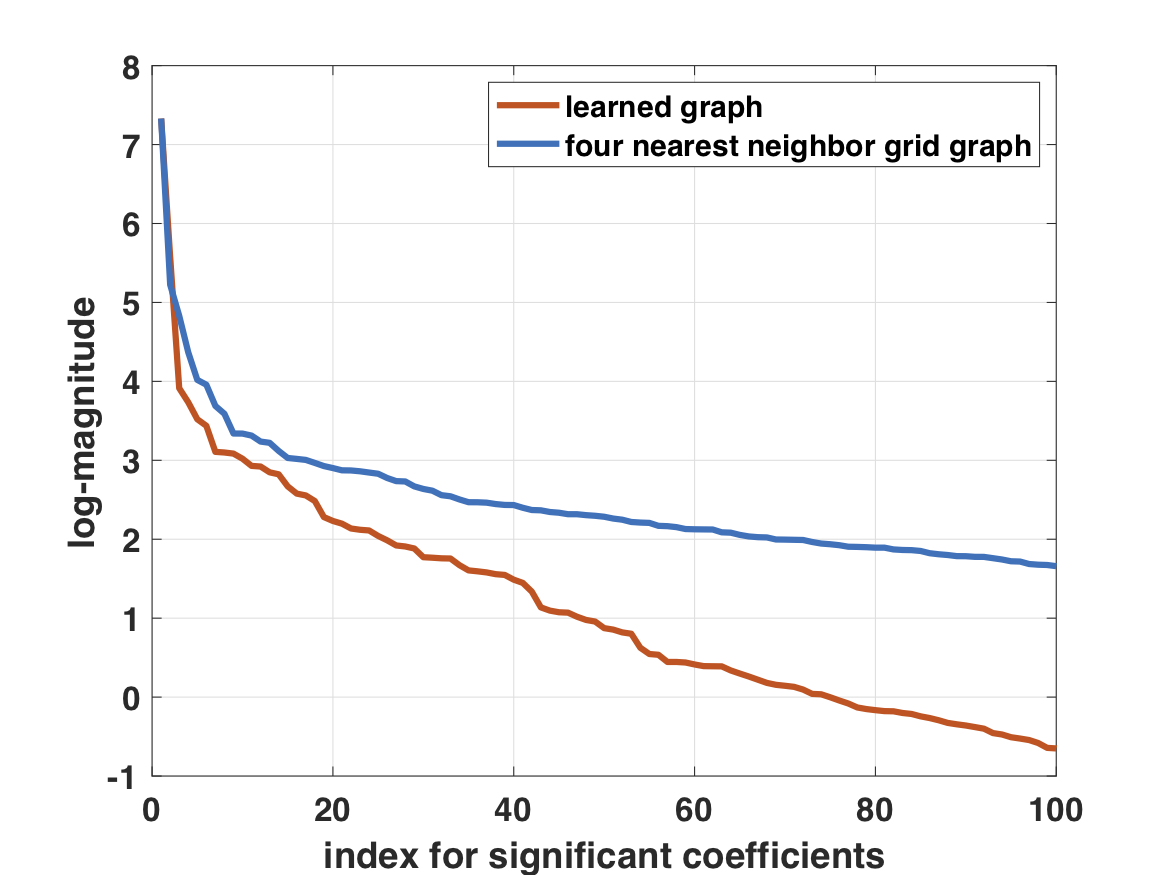}   
		\label{fig:GFT-decay}}
      \subfloat[]
		{ \includegraphics[width=7cm]{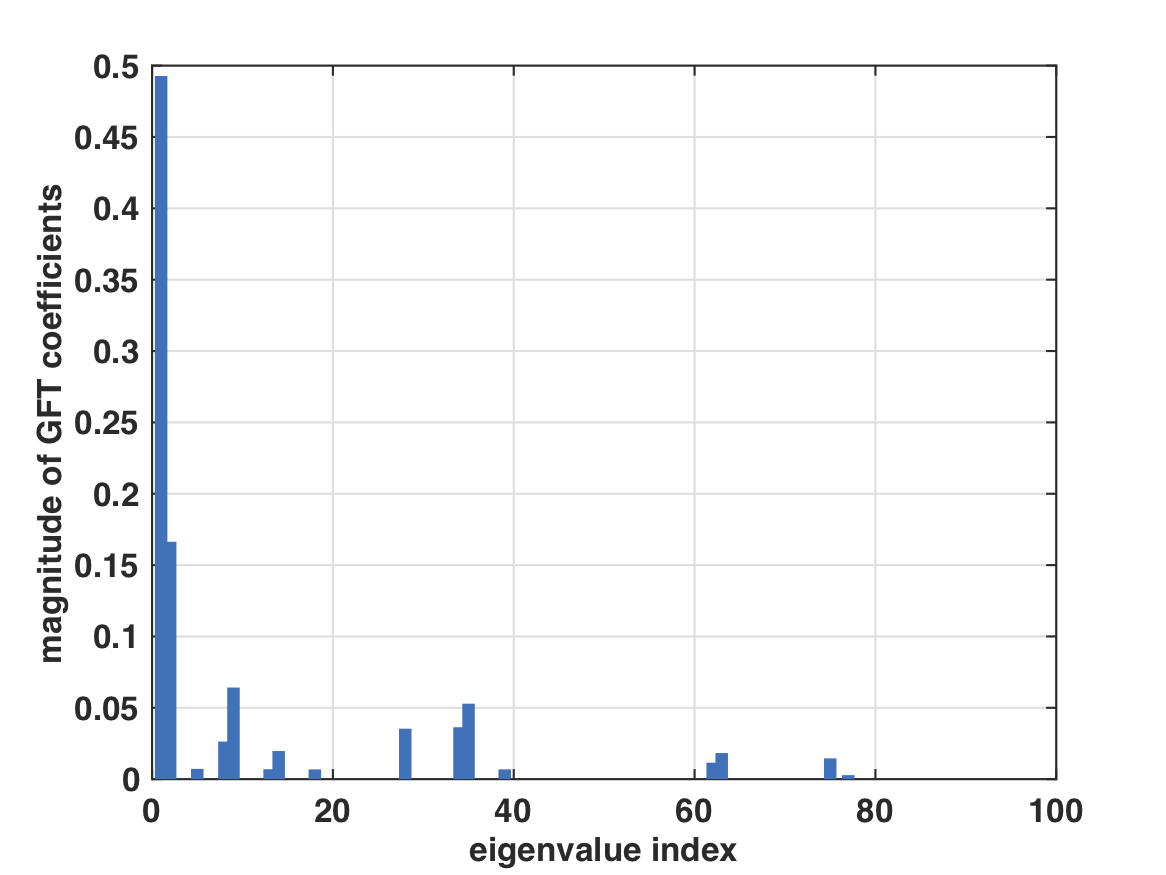}   
		\label{fig:GFT_image_compression}}
        \caption{{Inferring a graph for image coding: (a) The graph learned on a random patch of the image Teddy using \cite{Fracastoro2017}. (b) Comparison between the GFT coefficients of the image signal on the learned graph and the four nearest neighbor grid graph. The coefficients are ordered decreasingly by log-magnitude. (c) The GFT coefficients of the graph weights.}}
        \label{fig:compression-example}
\end{figure}

Graph learning has been introduced only recently for this type of problems. A learning model based on signal smoothness, inspired by \cite{Dong16,Kalofolias17}, has been further extended in order to design a graph-based coding framework that takes into account the coding of the signal values as well as the cost of transmitting the graph in rate distortion terms \cite{Fracastoro2017}.
In particular, the cost of coding the image signal is minimized by promoting its smoothness on the learned topology. The transmission cost of the graph itself is further controlled by adding an additional term in the optimization problem which penalizes the sparsity of the graph Fourier coefficients of the edge weight signal. 

{An illustrative example of the graph-based transform coding proposed in \cite{Fracastoro2017}, as well as its application to image compression, is shown in Fig.~\ref{fig:compression-example}. Briefly, the compression algorithm consists of three important parts. First, the solution to an optimization problem that takes into account the rate approximation of the image signal at a patch level, as well as the cost of transmitting the graph, provides a graph topology (Fig.~\ref{fig:compression-example}(a)) that defines the optimal coding strategy. Second, the GFT coefficients of the image signal on the learned graph can be used to compress efficiently the image. As we can see in Fig.~\ref{fig:compression-example}(b), the decay of these coefficients (in terms of their log-magnitude) is much faster than the decay of the GFT coefficients corresponding to a regular grid graph that does not involve any learning. Third, the weights of the learned graph are treated as a new edge weight signal that lies on a dual graph, whose nodes represent the edges in the learned graph, with the signal values on the nodes being the edge weights of the learned graph. Two nodes are connected in this dual graph if and only if the two corresponding edges share one common node in the learned graph. The learned graph is then transmitted by the GFT coefficients of this edge weight signal, where the decay of these coefficients is shown in Fig.~\ref{fig:compression-example}(c). The obtained results confirm that the GFT coefficients of the graph weights are concentrated on the low frequencies, which indicates a highly compressible graph.}

{Another example is the work in \cite{Lu2017} that introduces an efficient graph learning approach for fast graph Fourier transform that is based on \cite{Egilmez17}. The authors have considered a maximum likelihood estimation problem with additional constraints based on a matrix factorization of the graph Laplacian matrix, such that its eigenvector matrix is a product of a block diagonal matrix and a butterfly-like matrix. The learned graph leads to a fast non-separable transform for intra predictive residual blocks in video compression. Such efforts confirm that learning a meaningful graph can have a significant impact in graph-based image compression. These are only some first attempts which leave much room for improvement, especially in terms of coding performance. Thus, we expect to see more research efforts in the future to fully exploit the potential of graph methods.}

\subsection{Brain signal analysis}
GSP has been shown to be a promising and powerful framework for  brain network data, mainly due to the potential to jointly model the brain structure through the graph and the brain activity as a signal residing on the nodes of the graph.  The overview paper \cite{Huang18} provides a summary of how a graph signal processing view on brain signals can provide additional insights into the functionality of the brain. 

Graph learning in particular has been successfully applied for inferring the structural and functional connectivity of the brain related to different diseases or external stimuli. For example, \cite{Hu15} introduced a graph regression model for learning brain \emph{structural} connectivity of patients with Alzheimer's disease, which is based on the signal smoothness model discussed in Section \ref{sec:smoothness}. A similar framework \cite{Liu2018}, extended to the noisy settings, has been applied on a set of magnetoencephalography (MEG) signals to capture the brain activity in two categories of visual stimuli (e.g., the subject was viewing face or non-face images). In addition to the smoothness assumption, the proposed framework is based on the assumption that the perturbation on the low-rank components of the noisy signals is sparse. The recovered \emph{functional} connectivity graphs under these assumptions are compatible with findings in the neuroscientific literature, which is a promising result indicating that graph learning can contribute to this application domain. 

Instead of the smoothness model adopted in \cite{Hu15,Liu2018}, the authors in \cite{Shen16} have utilized models on causal dependencies and proposed to infer \emph{effective} connectivity networks of brain regions that may shed light on the understanding of the cause behind epilepsy. The signals that they use are electrocorticography (ECoG) time series data before and after ictal onset of seizures of epilepsy. All these applications show the potential impact GSP-based graph learning methods may have on brain and more generally biomedical data analysis where the inference of hidden functional connections can be crucial.

\subsection{Other application domains}
In addition to image processing and biomedical analysis, GSP-based graph learning methods have been applied to a number of other diverse areas. One notable example is meteorology, where it is of interest to understand the relationship between different locations based on the temperatures recorded at the weather stations in these locations. Interestingly, this is an area where all the three major signal models introduced in this tutorial may be employed to learn graphs that lead to different insights.
For instance, the authors of \cite{Dong16,Chepuri17} have proposed to learn a network of weather stations using the signal smoothness model, which essentially captures the relationship between these stations in terms of their altitude. Alternatively, the work in \cite{Pasdeloup18} has adopted the heat diffusion model in which the evolution of temperatures in different regions is modeled as a diffusion process on the learned geographical graph. The authors of \cite{Mei17} have further developed a framework based on causal dependencies to infer a directed temperature propagation network that is consistent with major wind directions over the United States. 
We note, however, that most of these studies are proof of concept, and future research is expected to focus more on the perspective of practical applications in meteorology.

Another area of interest is environmental monitoring. As an example, the author of \cite{Jablonski17} has proposed to apply the GSP-based graph learning framework of \cite{Kalofolias17} for the analysis of exemplary environmental data of ozone concentration in Poland. More specifically, the paper has proposed to learn a network that reflects the relationship between different regions in terms of ozone concentration. Such relationship may be understood in a dynamic fashion using data from different temporal periods.
Similarly, the work in \cite{Dong16} has analyzed evapotranspiration data collected in California to understand relationship between regions of different geological features.

Finally, GSP-based methods have also been applied to infer graphs that reveal urban traffic flows \cite{Thanou17}, patterns of news propagation on the Internet \cite{Baingana17}, inter-region political relationship \cite{Dong16}, similarity between animal species \cite{Egilmez17}, and ontologies of concepts \cite{Lake10}. The diversity of these areas has demonstrated the potential of applying GSP-based graph learning methods for understanding hidden relationship behind data observations in real world applications.

\section{Concluding remarks and future directions}
Learning structures and graphs from data observations is an important problem in modern data analytics, and the novel signal processing approaches reviewed in this paper have both theoretical and practical significance. 
On the one hand, GSP provides a new theoretical framework for graph learning by utilizing signal processing tools, with a strong emphasis on the representation of the signals on the learned graph, which can be essential from a modeling viewpoint. As a result, the novel approaches developed in this field would benefit not only the inference of optimal graph topologies, but potentially also the subsequent signal processing and data analysis tasks.
On the other hand, the novel signal and graph models designed from a GSP perspective may contribute uniquely to the understanding of the often complex data structure and generative processes of the observations made in real world application domains, such as brain and social network analysis. For these reasons, GSP-based approaches for graph learning have since recently attracted an increasing amount of interest; there exist, however, many open issues and questions that are worthy of further investigation. In what follows, we {discuss five general directions for future work.}

\subsection{Input signals of learning frameworks}
\label{sec:input}
The first important point that needs further investigation is the quality of the input signals.
Most of the approaches in the literature have focused on the scenario where a complete set of data is observed for all the entities of interest (i.e., at all vertices in the graph). However, there are often situations when observations are only partially available either due to failure in data acquisition from some sensors or simply because of the cost of making full observations. For example, in large-scale social, biomedical or environmental networks, sampling or active learning may need to be applied to select a limited number of sensors for observations \cite{Gadde14}. It is a challenge to design graph learning approaches that can handle such cases, and to study the extent to which the partial or missing observations affect the learning performance.
Another scenario is dealing with sequential input data that come in an online and adaptive fashion, which has been studied in the recent work of \cite{Vlaski18}.

\subsection{Outcome of learning frameworks}
\label{sec:output}
Compared to the input signals, it is perhaps even more important to rethink the potential outcome of the learning frameworks. Several important lines of thoughts remain largely unexplored in the current literature. First, while most of the existing work focuses on learning undirected graphs, it is certainly of interest to investigate approaches for learning directed ones. 
Methods described in Section~\ref{sec:causal}, such as \cite{Mei17,Baingana17,Shen16}, are able to achieve this since they do not explicitly rely on the notion of frequency provided by the eigendecomposition of the symmetric graph adjacency or Laplacian matrices. However, it is certainly possible and desirable to extend the frequency interpretation obtained with undirected graphs to the case of directed ones. For example, alternative definitions of frequencies of graph signals have been recently proposed based on normalization of the random walk Laplacian \cite{Mhaskar18}, novel definition of inner product of graph signals \cite{Girault18}, and explicit optimization for an orthonormal basis on graphs \cite{Sardellitti17,Shafipour18b}. {How to design techniques that learn directed graphs by making use of these new developments in the frequency interpretation of graph signals remains an interesting question.}

Second, in many real world applications, noticeably social network interactions and brain functional connectivities, the network structure changes over time. It is therefore interesting to look into learning frameworks that can infer dynamic graph topologies. To this end, \cite{Baingana17} proposes a method to track network structure that can be switched between a number of different states. Alternatively, \cite{Kalofolias17} has proposed to infer dynamic networks from observations within different time windows, with a penalty term imposed on the similarity between consecutive networks to be inferred.
Such a notion of temporal smoothness is certainly an interesting question to study, which may draw inspiration from visualizations of dynamic networks recently proposed in \cite{DalCol17}.

Third, although the current lines of work reviewed in this survey mainly focus on the signal representation, it is also possible to put constraints directly on the learned graphs by enforcing certain graph properties that go beyond the common choice of sparsity, which has been adopted explicitly in the optimization problems in many existing methods such as the ones in \cite{Friedman08,Lake10,Chepuri17,Pasdeloup18,Segarra17a,Mei17,Baingana17}.
One example is the work in \cite{Pavez18}, where the authors have proposed to infer graphs with monotone topology properties. Another example is the approach in \cite{Sudin17} which learns a sparse graph with connected components. Learning graphs with desirable properties inspired by a specific application domain (e.g., community detection \cite{Fortunato10}) can also have great potential benefit, and it is a topic worth investigating.

Fourth, in some applications it might not be necessary to learn the full graph topology, but some other intermediate or graph-related representations. For example, this can be an embedding of the vertices in the graph for the purpose of clustering \cite{Dong14}, or a function of the graph such as graph filters for the subsequent signal processing tasks \cite{Segarra17b}. Another possibility is to learn graph properties such as the eigenvalues (for example using technique described in \cite{Pasdeloup18}) or degree distribution, or templates that constitute local regions of the graph. Similar to the previous point, in these scenarios, the learning framework needs to be designed accordingly with the end objective or application in mind.

Finally, instead of learning a deterministic graph structure as in most existing methods, it would be interesting to explore the possibility of learning graphs in a probabilistic fashion in which we specify the confidence in building an edge between each pair of the vertices. This would benefit situations when a soft decision is preferred to a hard decision, possibly due to anticipated measurement errors in the observations or other constraints.

\subsection{Signal models}
Throughout this tutorial, we have emphasized the important role a properly defined signal model plays in the design of the graph learning framework. The current literature predominantly focuses on either the globally or locally smooth models. Other models such as bandlimited signals, i.e., the ones that have limited support in the graph spectral domain, may also be considered for inferring graph topologies \cite{Sardellitti16}. More generally, more flexible signal models that go beyond the smoothness-based criteria can be designed by taking into account general filtering operations of signals on the graph. 

The learning framework may also need to adapt to the specific input and output as outlined in Section~\ref{sec:input} and Section~\ref{sec:output}. 
For instance, given only partially available observations, it might make sense to consider a signal model tailored for the observed, instead of the whole, region of the graph. Another scenario would be that, in learning dynamic graph topologies, the signal model employed needs to be consistent with the temporal smoothness criteria adopted to learn the sequence of graphs.

\subsection{Performance guarantees}
Graph inference is an inherently difficult problem given the large number of unknown variables (generally in the order of $N^2$) and the relatively small amount of observations. As a result, learning algorithms need to be designed with additional assumptions or priors. In this case, it is desirable to have theoretical guarantees on the performance of graph recovery under the specific model and prior. 
It would also be interesting to put the errors in graph recovery into the context of the subsequent data processing tasks and study their impact.
Furthermore, for many graph learning algorithms, in addition to the empirical performance it is necessary to provide guarantees of the convergence when alternating minimization is employed, as well as to study the computational complexity that can be essential for learning large-scale graphs. These theoretical considerations remain largely unexplored in the current literature and hence require much further investigation, given their importance.

\subsection{Objective of graph learning}
{The final comment on future work is a reflection on the objective of the graph learning problem and, in particular, how to better integrate the inference framework with the subsequent data analysis tasks. Clearly, the learned graph may be readily used for classical machine learning tasks such as clustering or semi-supervised learning, but it may also directly benefit the processing and analysis of the graph signals.
In this setting, it is often the case that a cost related to the application is directly incorporated into the optimization for graph learning. For instance, the work in \cite{Yankelevsky16} has proposed a method for inferring graph topologies with a joint goal of dictionary learning, whose cost function is incorporated into the optimization problem. In many applications, such as image coding, accuracy is not the only interesting performance metric. Typically, there exist different trade-offs that are more complex and should be taken into consideration. For example, in image compression, the actual cost of coding the graph is at least equally important compared to the cost of coding the image signal. Such constraints are indicated by the application, and they should be incorporated in the graph learning framework (e.g., \cite{Fracastoro2017}) in order to make the learning framework more targeted to a specific application.}

\section{Acknowledgements}
The authors would like to thank Giulia Fracastoro for her help with preparing Fig.~\ref{fig:compression-example}.

\bibliographystyle{IEEEtran.bst}
\bibliography{mybibfile.bib}

\end{document}